\documentclass[10pt,twocolumn,letterpaper]{article}

\usepackage[pagenumbers]{cvpr} % To force page numbers, e.g. for an arXiv version

\usepackage{amsmath} % For arrows
\usepackage{multirow} % For multirow
\usepackage{booktabs}
\usepackage{colortbl}
\usepackage{makecell}
\usepackage{pifont}% http://ctan.org/pkg/pifont
\usepackage{dblfloatfix}
\usepackage{tabularx}
\usepackage{placeins}
\usepackage{pifont}
\newcommand{\cmark}{\ding{51}}%
\newcommand{\xmark}{\ding{55}}%
\usepackage{caption}
\usepackage{makecell} % Enables line breaks in cells
\usepackage[accsupp]{axessibility}
\colorlet{lightred}{Red!35}
\colorlet{lightorange}{Dandelion!50}
\colorlet{lightyellow}{Yellow!30}
\colorlet{lighttop1}{Dandelion!50}
\colorlet{lighttop2}{Yellow!30}
\colorlet{rowhighlight}{Blue!13}
\colorlet{columnhighlight}{Gray!15}

\definecolor{cvprblue}{rgb}{0.21,0.49,0.74}
\usepackage[pagebackref,breaklinks,colorlinks,allcolors=cvprblue]{hyperref}

\def\paperID{xxxx} % *** Enter the Paper ID here
\def\confName{CVPR}
\def\confYear{2025}

\title{ReCap: Better Gaussian Relighting with Cross-Environment Captures}

\author{Jingzhi Li \hspace{0.2cm} Zongwei Wu\textsuperscript{*} \hspace{0.2cm} Eduard Zamfir \hspace{0.2cm} Radu Timofte\\
{\normalsize Computer Vision Lab, CAIDAS \& IFI, University of Würzburg}\\
}

\begin{document}
\twocolumn[{
\renewcommand\twocolumn[1][]{#1}%
\maketitle
\begin{center}
   \includegraphics[width=1\textwidth,height=6cm,keepaspectratio]{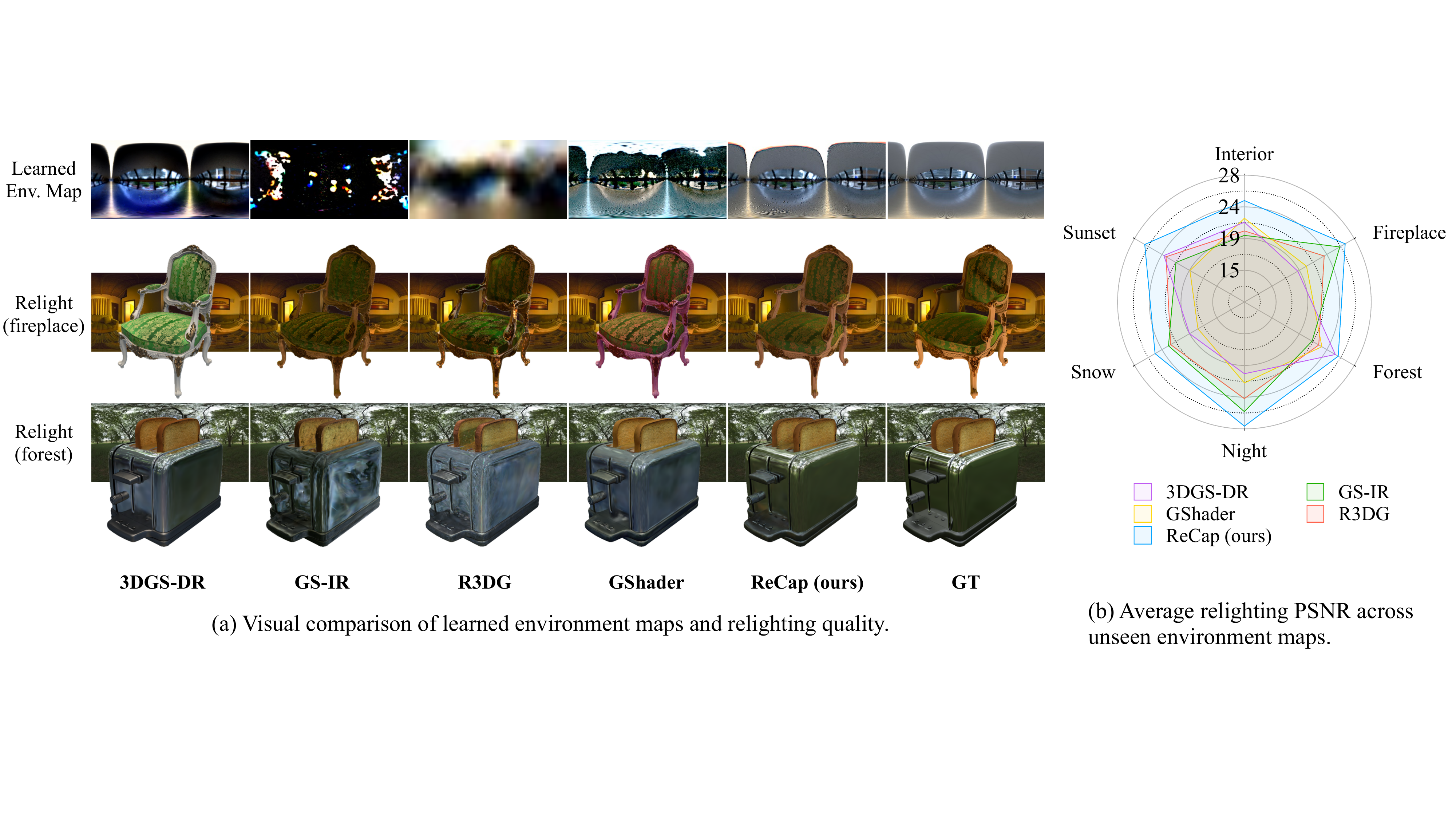}
   \vspace{-0.5em}
    \captionof{figure}{ReCap shows a significant advantage in reconstructing environment maps with accurate tones and color fidelity. Both qualitative and quantitative assessments show that ReCap achieves more realistic and consistent relighting results under a range of unseen lighting conditions. }
    \label{fig:teaser}
\end{center}
}]
\begin{abstract}
\renewcommand{\thefootnote}{\fnsymbol{footnote}}
\footnotetext[1]{Corresponding author}
Accurate 3D objects relighting in diverse unseen environments is crucial for realistic virtual object placement. Due to the albedo-lighting ambiguity, existing methods often fall short in producing faithful relights. Without proper constraints, observed training views can be explained by numerous combinations of lighting and material attributes, lacking physical correspondence with the actual environment maps used for relighting. In this work, we present ReCap, treating cross-environment captures as multi-task target to provide the missing supervision that cuts through the entanglement. Specifically, ReCap jointly optimizes multiple lighting representations that share a common set of material attributes. This naturally harmonizes a coherent set of lighting representations around the mutual material attributes, exploiting commonalities and differences across varied object appearances. 
Such coherence enables physically sound lighting reconstruction and robust material estimation — both essential for accurate relighting. Together with a streamlined shading function and effective post-processing, ReCap outperforms all leading competitors on an expanded relighting benchmark.
\end{abstract}
\section{Introduction}

\label{sec:intro}
For a realistic and immersive augmented reality experience, virtually placed objects must convincingly reflect light, cast shadows, and adapt naturally to different lighting conditions. Achieving the desired realism requires a physically accurate response to environment lighting, driving a line of research at enabling relighting capabilities in popular neural representation models.

In recent years, Neural Radiance Field (NeRF)~\cite{mildenhall2020nerf} gained prevailing popularity as an \textit{implicit} scene representation. Subsequent NeRF-based relighting methods~\cite{srinivasan2021nerv,zhang2021nerfactor,jin2023tensoir} produced impressive relighting results, but their computational demands make them impractical for interactive applications~\cite{jiang2024gaussianshader,ye20243d}. Lately, 3D Gaussian Splatting (3DGS)~\cite{kerbl20233d} is widely acclaimed as an \textit{explicit} 3D representation model for its high rendering quality and interactive frame rates, naturally suited for applications requiring real-time performance. Building on this, follow-up works~\cite{jiang2024gaussianshader,liang2024gs,ye20243d,R3DG2023} have enabled relighting of Gaussians using explicit shading functions and learnable lighting representations, often in the form of environment maps. 

While standard HDR maps could theoretically replace these learned environment maps for relighting, directly substituting them, as current methods do, remains questionable due to the unclear physical meaning of the learned values. Because of the \textit{albedo-lighting ambiguity}~\cite{barron2012shape,barrow1978recovering}, where changes in surface albedo are indistinguishable from changes in lighting intensity, existing supervision from the reconstruction loss alone is not enough for a truthful lighting reconstruction. As shown in~\cref{fig:teaser}a), the learned environments are often observed to be tinted with object colors, shifted in tone, scaled in intensity or filled with noise. Without proper constraints, these maps act as sinks for unmodeled residual terms during optimization, becoming indispensable for producing high-quality novel view synthesis results. Replacing them with the ground truth HDR maps significantly degrades the output quality, much less when attempting to relight with novel environment maps.

Inspired by photometric appearance modeling~\cite{pintus2019state,li2020mvps,xu2019deep} which estimates surface properties from object appearances under varied lighting, we introduce additional photometric supervision to address the albedo-lighting ambiguity. While traditional approaches rely on controlled lighting setups like light stages~\cite{debevec2012light} and collocated lights~\cite{xu2018deep} to provide supervision through known light directions and/or intensities, they require dedicated hardwares. Instead, we propose ReCap to leverage object captures across unknown lighting conditions, modeling light-dependent appearances with multiple environment maps that share a common Gaussian model. Conceptually, this resembles multi-task learning, where the learned environment maps act as task heads querying a shared material representation for varied object appearances. In this case, the ``querying" is done by the physically-based shading function, which facilitates the separation of material and lighting. This joint optimization promotes internal consistency when accounting for varied object appearances across diverse environments, simultaneously supporting more accurate lighting reconstruction.

To streamline the joint optimization, we introduce a generalized shading function based on the split-sum approximation~\cite{karis2013real}, which eliminates a material parameter that introduces ambiguity in inverse rendering. Additionally, we ensure compatibility with standard HDR maps by encouraging the learned values to remain in a linear HDR space through appropriate post-processing. This enables the direct application of novel environment maps without the need for image normalization~\cite{liu2023nero} or map adjustments~\cite{liang2024gs}.

Current relighting evaluations are limited in scope, focusing primarily on diffuse surfaces~\cite{jin2023tensoir,zhang2022modeling}. For a more comprehensive assessment, we re-rendered $13$ objects from NeRF~\cite{mildenhall2020nerf} and RefNeRF~\cite{verbin2022ref} featuring both diffuse and specular surfaces. As shown in~\cref{fig:teaser}, experiments on the expanded benchmark confirm the effectiveness of ReCap in producing robust and more realistic relighting results. Codes and datasets are available \href{https://jingzhi.github.io/ReCap/}{here}.

To summarize, the contribution of this work includes
\begin{itemize}
    \item Treating cross-environment captures as multi-task targets, we address the albedo-lighting ambiguity via the joint optimization of shared material properties and independent lighting representations.
    \item We propose a novel shading function with physically appropriate post-processing, providing more flexible material representation that eases optimization and allows the direct application of standard HDR maps.
    \item ReCap achieves state-of-the-art relighting performance on a more comprehensive benchmark, showing robustness across diverse lighting and object types.
\end{itemize}
\begin{figure*}[ht]
    \centering  
    \includegraphics[width=0.9\linewidth]{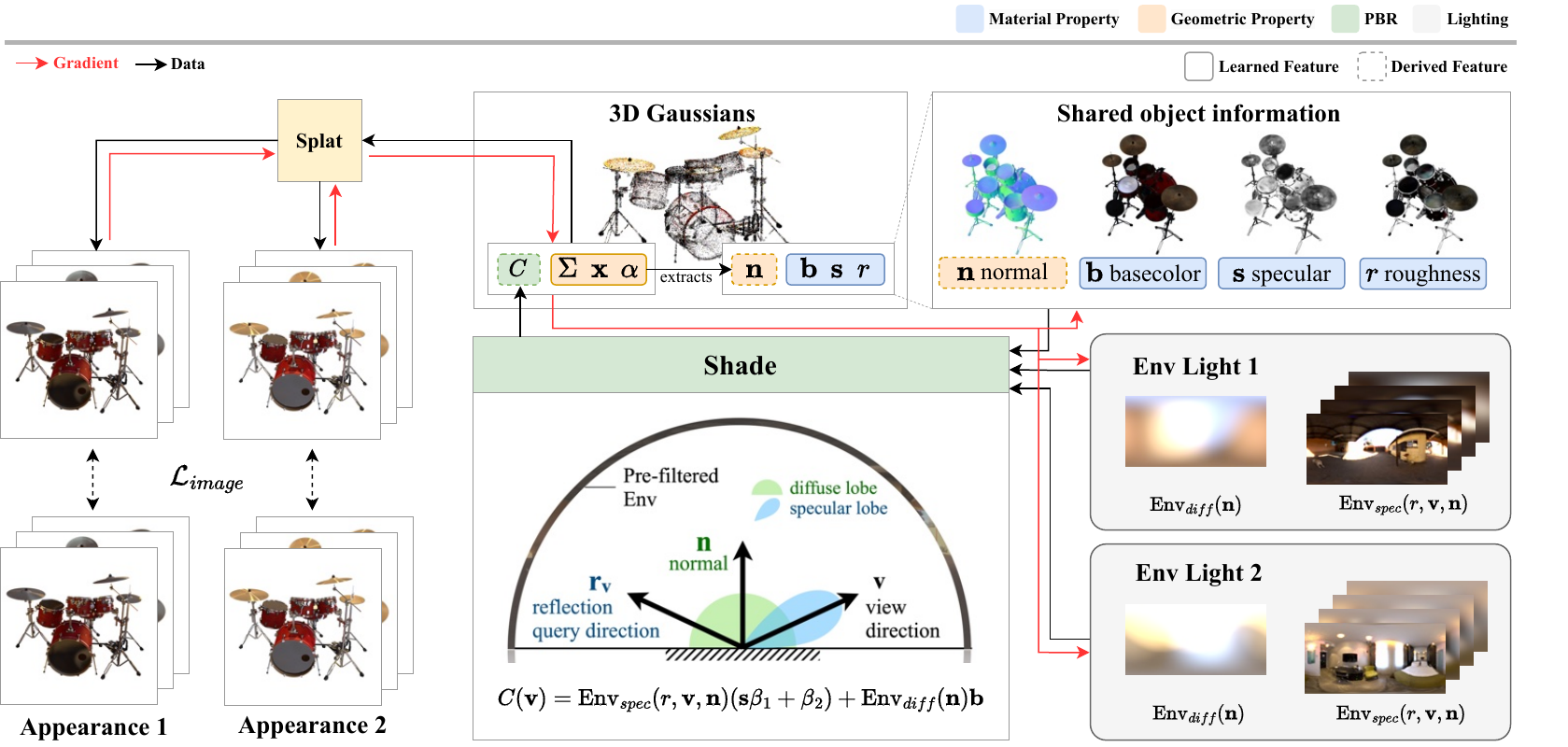}
    \scriptsize
    \vspace{-1em}
    \caption{\textbf{The proposed ReCap training framework}. Compared to original 3DGS~\cite{kerbl20233d}, each Gaussian is augmented with 3 extra material attributes.  Given $k$ sets of object appearances from unknown lighting conditions as input, $k$ learnable environment maps are instantiated. Gaussian color is computed according to the shading function in the world space based on environment queries and material properties.  2D images are rasterized with standard Gaussian splatting and used for loss computation. $\mathcal{L}_{\text{image}}$: image reconstruction loss from \cite{kerbl20233d}. Additional loss terms for material and geometry are not shown.}
    \label{fig:method}
    \vspace{-2em}
\end{figure*}
\section{Related Work}
\label{sec:related}
\noindent{\textbf{Novel View Synthesis (NVS).}
Pure NVS techniques focus on reproducing the scene appearance under its original environment. NeRF~\cite{mildenhall2020nerf} and its follow-ups~\cite{barron2021mip,barron2022mip,barron2023zip} achieve remarkable NVS quality with implicit radiance fields and volume rendering. Although great progress has been made in NeRF acceleration~\cite{fridovich2022plenoxels,muller2022instant,xu2022point} leveraging voxel grids or hash tables, their rendering speeds ($\sim$10 fps) are still far from interactive. Building on progress in differentiable point-based rendering~\cite{yifan2019differentiable,kopanas2021point}, 3DGS~\cite{kerbl20233d} recently introduced an explicit Gaussian representation with an efficient tile-based rasterizer, delivering photo-realistic rendering at impressive frame rates ($\sim$100 fps). This makes 3DGS well-suited for high-quality real-time relighting.

\noindent{\textbf{Relighting.}}
The relighting task traditionally involves editing lighting within fixed views~\cite{debevec2000acquiring,ren2015image,xu2018deep}. The term has since broadened to encompass novel view relighting~\cite{xu2019deep, kang2019learning}, which aims to generate images from both new viewpoints and under new lighting conditions. With the advent of differentiable neural rendering, reconstruction-based methods~\cite{boss2021nerd,zhang2021physg,zhang2021nerfactor,srinivasan2021nerv} have found a promising direction to combine physically-based rendering (PBR) with neural representations for editable and realistic rendering. Closely related to our work, PBR is also incorporated by 3D Gaussians, which enable Gaussian relighting. 3DGS-DR~\cite{ye20243d} and GShader~\cite{jiang2024gaussianshader} both specialize in specular objects and propose customized shading functions. R3DG~\cite{R3DG2023} further incorporates ray tracing for indirect illumination, while GS-IR~\cite{liang2024gs} relies on baked occlusion maps. With only single environment captures as input, these Gaussian relighting methods face challenges in correctly decoupling lighting from material properties.

\noindent{\textbf{Albedo-lighting Disambiguation.}}
Recovering reliable illumination and albedo from object appearances is an ill-posed problem~\cite{ramamoorthi2001signal} in inverse rendering. It is often simplified under the assumption of controllable lighting~\cite{bi2020deep,nam2018practical}, known geometry~\cite{park2020seeing}, or when regularized with strong priors~\cite{ren2023improving,feng2022towards} in domain-specific tasks such as relighting of human faces. For more general cases, object captures from varied lighting, known as photometric images, can provide important visual cues. Several NeRF-based relighting methods have taken advantage of such captures. NeRV~\cite{srinivasan2021nerv} requires multiple known lighting environments. NeRD~\cite{boss2021nerd} models lighting as spherical Gaussians and optimizes separate lighting for every input image. TensoIR~\cite{jin2023tensoir} encodes lighting as an additional dimension in its factorized tensor representation and relies on ground-truth albedo scaling for accurate relighting. In contrast, we use object captures under unknown lighting, optimizing a learnable light map separately for each scene and independently from the Gaussian model, without requiring ground-truth albedo for relighting.

\vspace{-1em}
\section{Method}
\setlength{\abovedisplayskip}{4pt}    % Default is 10pt
\setlength{\belowdisplayskip}{4pt}    % Default is 10pt
\label{sec:method}
In this section, we introduce ReCap, a robust Gaussian relighting method that leverages cross-environment object captures. The overall framework is illustrated in \cref{fig:method}. In \cref{sec:relight_w_shading}, we explain how relighting is enabled in 3DGS with an explicit shading function. In \cref{sec:split_sum}, an optimization-friendly variant of the split-sum-approximation is introduced for shading. Lighting representation and post-processing are detailed in \cref{sec:lighting} and \cref{sec:postprocess}, with normal estimation covered in \cref{sec:geometry}.
\vspace{-1em}
\subsection{Relighting Gaussians with shading function}
\label{sec:relight_w_shading}
3DGS represents objects as explicit point clouds. A point at location $\mathbf{x}$ in the world space is represented as a 3D Gaussian function defined by covariance matrix $\Sigma$ and mean $\mu$:
\begin{equation}
    \mathcal{G}(\mathbf{x}|\mu,\Sigma)=e^{\frac{1}{2}(\mathbf{x}-\mu)^T\Sigma^{-1}(\mathbf{x}-\mu)}
\end{equation}
Additionally, each point holds an opacity attribute $\alpha$ and a color attribute $\mathbf{c}$ for point-based $\alpha$-blending. From a viewing direction $\mathbf{v}$, the pixel color $C$ can be computed by blending ordered points overlapping the pixel: 
\begin{equation}
C(\mathbf{v})=\sum_i\mathbf{c}_i(\mathbf{v})\alpha_i\prod_{j=1}^{j-1}(1-\alpha_j)
\end{equation}
The original 3DGS models the view-dependent color $\mathbf{c}(\mathbf{v})$ with spherical harmonics. This simplification abstracts the complex view-dependent interactions between material, lighting and geometry into a composite representation. Relighting Gaussians thus requires an alternative representation of $\mathbf{c}(\mathbf{v})$ that factors out the influence of lighting. A natural choice is to use various well-established shading functions from graphics. 
Ignoring any emission, let the classic rendering equation represent the outgoing radiance at Gaussian point $\mathbf{x}$ viewed from direction $\mathbf{v}$ as
\begin{equation}
L_{out}(\mathbf{x}, \mathbf{v}) =\int_\Omega f_r(\mathbf{x},\mathbf{v},\mathbf{l})L_{in}(\mathbf{x}, \mathbf{l})
(\mathbf{n}\cdot\mathbf{l}) d\mathbf{l},
\label{eq:rendering_eq}
\end{equation}
where $\Omega$ is the hemisphere above the surface, $f_r$ is the bi-directional reflection function (BRDF), $\mathbf{l}$ is the incident light direction and $L_{in}$ is the incoming radiance, all defined in the local coordinate system centered at $\mathbf{x}$. The Gaussian color $\mathbf{c}$ can be computed from $L_{out}$ with proper post-processing such as tone mapping and gamma correction as discussed in~\cref{sec:postprocess}. The pixel color $C$ becomes
\begin{equation}
C(\mathbf{v})=\sum_i\mathbf{c}_i(\mathbf{v}|f_r,L_{in},\mathbf{n})\alpha_i\prod_{j=1}^{j-1}(1-\alpha_j).
\end{equation}

The rendering equation is commonly simplified for implementation as shading functions. GShader \cite{jiang2024gaussianshader} and GS-DR \cite{ye20243d} handcrafted their shading functions with a light-independent diffuse component and a light-dependent specular component querying lighting information from a learned environment map. R3DG~\cite{R3DG2023} and GS-IR~\cite{liang2024gs} both leverage microfacet based model as a more expressive alternative. In our work, we start with the split-sum approximation of the microfacet model and propose a more optimization-friendly variant.
\subsection{Disambiguate Split Sum Approximation}
\label{sec:split_sum}
Disney Principled BRDF~\cite{burley2012physically} provides user-friendly parameters building on the Cook-Torrance microfacet BRDF~\cite{cook1981reflectance}. We adopt simplifications from Epic Game~\cite{karis2013real} and consider the following parameters: 1) \textit{basecolor}, $\mathbf{b}\in[0,1]^3$; 2) \textit{roughness}, $r\in[0,1]$; 3) \textit{metallic}, $m\in[0,1]$ ; and 4) \textit{specular}, $s = 0.04$, assumed to be constant for non-metals. 

The BRDF of interest is given as
\begin{equation}
\begin{split}
    f_r(\mathbf{x},\mathbf{v},\mathbf{l})=&(1-m)\frac{\mathbf{b}}{\pi}\\  &+\frac{D(r,\mathbf{n},\mathbf{l},\mathbf{v})F(\mathbf{b},m,s,\mathbf{l},\mathbf{v})G(r,\mathbf{n},\mathbf{l},\mathbf{v})}{4(\mathbf{l}\cdot\mathbf{n})(\mathbf{v}\cdot\mathbf{n})},
\end{split}
\label{eq:microfacet_brdf}
\end{equation}
where $D$, $F$ and $G$ are the normal distribution function, the Fresnel term and the geometry term respectively.% Refer to the supplementary for detailed explanation.

Substitute~\cref{eq:microfacet_brdf} into~\cref{eq:rendering_eq} and apply the split-sum approximation as described in ~\cite{karis2013real}, the shading function can be written as 
\begin{equation}
\begin{split}  
    L_{out}(\mathbf{x},\mathbf{v})&=\underbrace{E_{d}(\mathbf{n})(1-m)\mathbf{b}}_{diffuse}\\
    &+\underbrace{E_{s}(\mathbf{n},\mathbf{v})\left[F_0\beta_{1}(r,\mathbf{n},\mathbf{v})+\beta_{2}(r,\mathbf{n},\mathbf{v})\right]}_{specular},\\
\end{split}
\label{eq:spa_original}
\end{equation}
where $E_d$ and $E_s$ are the pre-filtered environment maps for diffuse and specular reflectance, $\beta_1$ and $\beta_2$ are pre-calculated BRDF look-ups, $F_0 =m\mathbf{b}+(1-m)s$ is the effective reflectance.

This is a linear blend of two different models controlled by the metallic parameter. For metals, there is no diffuse component and the specular part is colored. For non-metals, the diffuse part is colored but not the specular part. Dropping arguments for conciseness, the two models are
\begin{equation}
\begin{split}  
    L_{\text{metal}}&=E_s\mathbf{b}\beta_1+E_s\beta_2\\
    L_{\text{non-metal}}&= E_ss\beta_1+E_s\beta_2+E_d\mathbf{b}.\\
\end{split}
\label{eq:spa}
\end{equation}
Without prior knowledge on the distribution of material parameters of real metal and non-metal, optimization of the metallic parameter is problematic especially when the lighting is also learned. First, $s$ and $\mathbf{b}$ have overlaps. Certain specular highlights of dark surfaces may be misinterpreted and assigned to the wrong metallic value, as illustrated by the example in \cref{fig:metallic}. Furthermore, when the environment maps and roughness values are optimized such that $E_d\sim E_s\beta_1$, the two equations becomes interchangeable with some value of $\mathbf{b}$ leading to ambiguous optimization.

Instead, we discard the metallic parameter and expand the specular parameter to propose a general expression as
\begin{equation}
    L_{\text{out}}= E_s\mathbf{s}\beta_1+E_s\beta_2+E_d\mathbf{b},
\label{eq:ownshading}
\end{equation}
where $\mathbf{s}\in[0,1]^3$ is now a vector representing \textit{specular tint}. For $\mathbf{s}=[s,s,s]^T$, it becomes $ L_{\text{non-metal}}$; for $\mathbf{s}=\mathbf{b}_{\text{metal}}$ and $\mathbf{b}=0$, it becomes $L_{\text{metal}}$. On the downside, the expanded range of $\mathbf{s}$ encompasses a significant portion of unnatural specular colors. To avoid overly saturated specular tint, we apply saturation penalty on $\mathbf{s}$ as
\begin{equation}
    \mathcal{L}_{\text{sat}} = \lambda_{\text{sat}} \cdot \left\| \mathbf{s} - \mathbf{s}_{\text{mean}} \right\|.
\end{equation}
To account for energy conservation, we further include a regularizer to encourage $\left\|\mathbf{s}\right\|+\left\|\mathbf{b}\right\|\leq1$.
\begin{figure}
    \centering
    \includegraphics[width=0.8\linewidth]{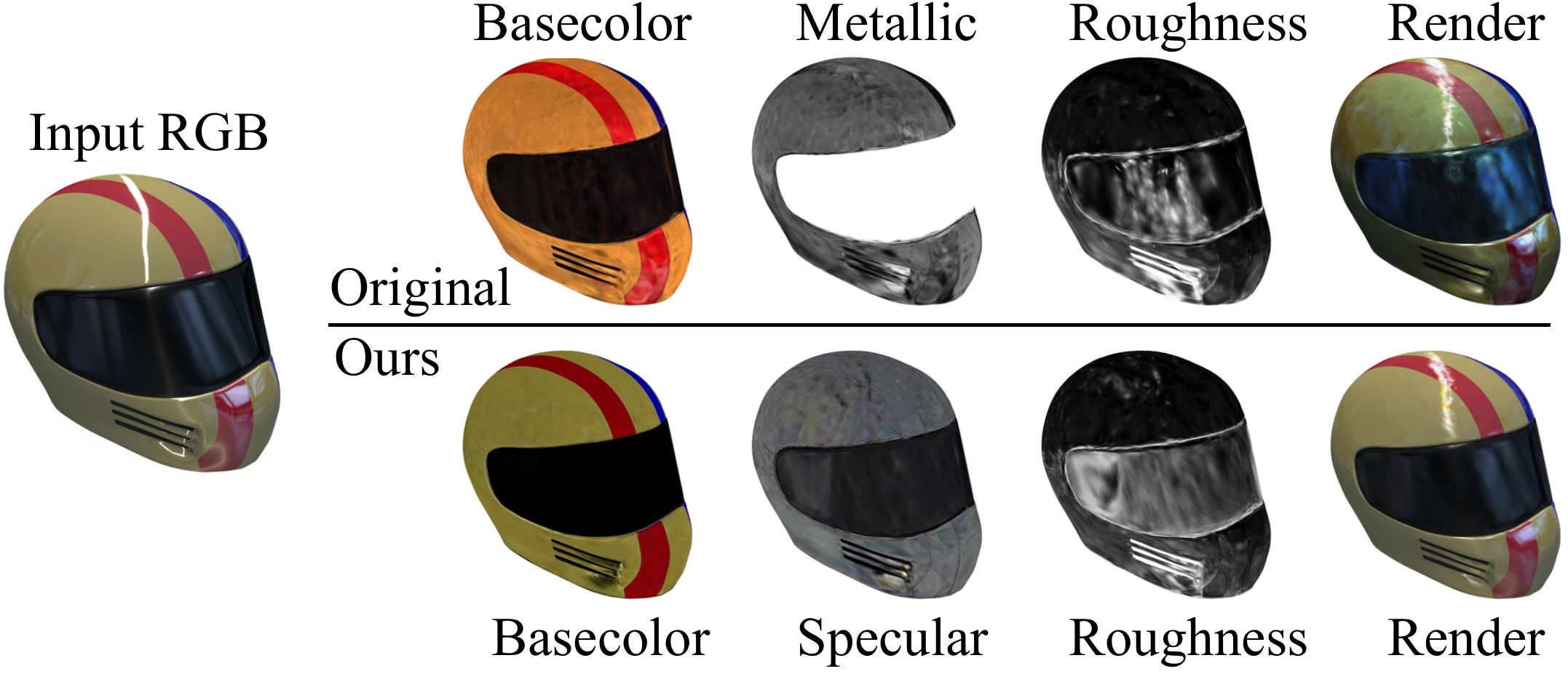}
    \vspace{-1em}
    \caption{With the original shading model, the shield of the helmet is falsely identified as being metallic during optimization.}
    \label{fig:metallic}
    \vspace{-2em}
\end{figure}
\subsection{Lighting Representation}
\label{sec:lighting}
While spherical functions~\cite{wang2009all,zhang2021physg,garon2019fast} are popular choices for efficient lighting representations, we adopt an image based lighting~\cite{debevec1998rendering} model to provide high frequency details necessary for specular reflections. Specifically, each environment lighting is represented by a $6\times256\times256$ learnable cube map, which is pre-filtered into a diffuse map, $E_d$, for diffuse reflection~\cite{ramamoorthi2001efficient} and a set of specular mipmaps across different roughness levels, $E_s$, for specular reflection~\cite{karis2013real}.  In practice, we use the efficient approximation provided by NVDIFFRAST~\cite{munkberg2022extracting} for differentiable pre-filtering and querying of the learnable environment maps, which are performed in every forward pass for shading.

To leverage photometric supervision from cross-environment captures, $k$ sets of learnable environment maps will be instantiated to explain $k$ sets for object appearances. As illustrated by~\cref{fig:env_illus}, the same object position can display a range of pixel colors depending on viewing direction and lighting. With only single-environment captures, the view-dependent variations can be explained by a multitude of material-lighting combinations, manifesting the albedo-lighting ambiguity. By introducing additional sets of object appearances under new, unknown environments, the light-dependent variations are attributed to corresponding learnable environment maps as multi-task learning targets. The joint optimization encourages physically sound decoupling of material properties and lighting to explain all observed appearances as guided by the shading function. 

While increasing the number of environments can enhance the decoupling as later shown in~\cref{sec:ablate}, it also requires additional effort to capture real objects across multiple scenes. As a proof of concept, we limit our investigation to a dual-environment setup unless mentioned otherwise. 
\begin{figure}
    \centering
    \includegraphics[width=1.0\linewidth]{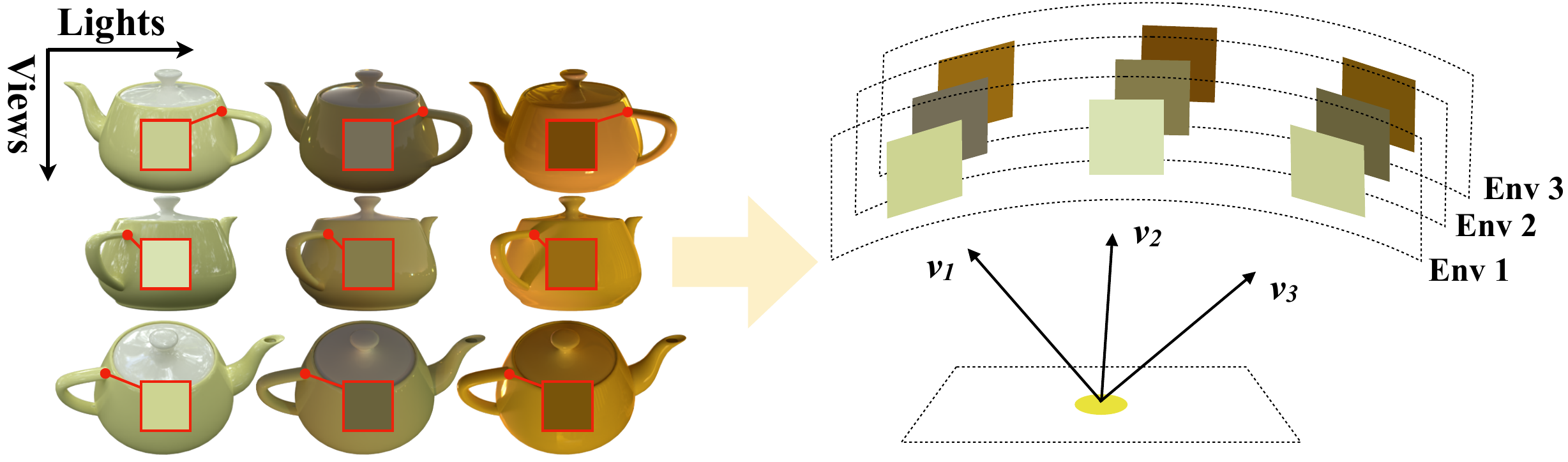}
    \caption{The same object position exhibit view-dependent and light-dependent pixel color, the later is accounted for by querying corresponding learnable environments. }
    \vspace{-2em}
    \label{fig:env_illus}
\end{figure}
\subsection{Post-shading Processing}
\label{sec:postprocess}
Standard HDR maps represent radiance in linear RGB space when used as lighting sources in rendering~\cite{debevec1998rendering}. To apply them effectively for relighting, the learned lighting values need a clear physical interpretation. Current methods adopt diverse post-shading strategies during training, influencing both the interpretation of learned environment maps and the processing of new HDR maps. For example, if the learned values are constrained to $[0,1]$ and gamma correction is applied post-shading, the environment map is effectively treated as being in a linear low dynamic range (LDR)~space.

In~\cref{tab:env_interpretation}, we analyze three major factors that affects relighting. Gamma correction is essential for relighting, as expected, since the shading function models linear light transport. However, it is often overlooked~\cite{R3DG2023,ye20243d}, because omitting it still yields good NVS results. While previous work also consider complex tone-mappers~\cite{srinivasan2021nerv} from graphics, they perform worse than simple clipping, likely due to the introduced non-linearity hindering optimization. 

To balance relighting and NVS, we constrain environment maps to be positive and apply simple clipping followed by gamma correction. This implies a linear HDR lighting space where new environment maps can be used without modification, removing the need for normalizing relit images~\cite{liu2023nero} or learned albedo~\cite{jin2023tensoir} to match the ground-truth (unavailable in practice) when when evaluating relighting. Combined with cross-environment decoupling, this essentially improves relighting by (i) allowing Gaussians to push light-dependent appearances into learnable ``sinks" as pseudo light maps, and (ii) regulating these ``sinks" with HDR processing and PBR, enabling them to approximate real ambient light in value distribution.
\begin{table}[ht]
    \centering
    \footnotesize
    \caption{Relighting and NVS performance (PSNR) of various design choices. The $\rightarrow$ symbol denotes implication on the learned environment map. During relighting, the input HDR map are pre-processed to match the interpretation. See the supplementary for more details.}
    \vspace{-1em}
    \resizebox{\linewidth}{!}{
    \begin{tabular}{ccc|cc}
         \toprule
         Env. Map Range &Tonemap&Gamma&Relight&NVS \\
         \hline
         $[0, 1]\rightarrow \text{LDR}$& \xmark &\xmark$\rightarrow \text{non-linear}$&23.55&29.97\\
         $[0, 1]\rightarrow \text{LDR}$&\xmark &\cmark$\rightarrow \text{linear}$&24.07&30.09\\
         $[0, \infty)\rightarrow \text{HDR}$& clip &\xmark$\rightarrow \text{non-linear}$&22.69&\textbf{32.36}\\
         $[0, \infty)\rightarrow \text{HDR}$& clip &\cmark$\rightarrow \text{linear}$ &\textbf{25.82}&32.23\\
         $[0, \infty)\rightarrow \text{HDR}$& reinhard &\cmark$\rightarrow \text{linear}$ &23.13&29.79\\
         $[0, \infty)\rightarrow \text{HDR}$& aces &\cmark$\rightarrow \text{linear}$ &23.70&30.64\\
         \bottomrule
    \end{tabular}}
    \label{tab:env_interpretation}
    \vspace{-2em}
\end{table}
\subsection{Geometry Estimation}
\label{sec:geometry}
Accurate normal estimation is essential for precise querying of the high frequency light maps. As discrete sparse point clouds, 3D Gaussians have no native support for normal computation. Following the common observation that converged Gaussians are often flat and align with the object surface~\cite{liang2024gs,jiang2024gaussianshader,guedon2024sugar,ye20243d}, we adopt the shortest axis of each Gaussian as an estimation of the normal vector. 
This shortest-axis normal approximation, $\mathbf{n}$, is simply constrained using depth-derived normal, $\mathbf{\hat{n}}$, where the depth image is rendered from Gaussian opacities. This depth-normal consistency loss is given as
\begin{equation}
    \mathcal{L}_{\text{dn}} = \lambda_{\text{dn}} \cdot \left\| \mathbf{n} - \mathbf{\hat{n}} \right\|^2.
\end{equation}
While existing works commonly augment normal estimation with additional residuals~\cite{jiang2024gaussianshader} or learn per-Gaussian normal vectors directly~\cite{R3DG2023,liang2024gs}, we observe that such learnable normals often overfit to explain specularities or shadowing effects. This leads to improved NVS but compromises geometric accuracy. As shown in~\cref{fig:normal}, the cross-lighting consistency imposed by ReCap naturally improves normal estimation by preventing overfitting to a single lighting condition. When only single-environment captures are used, specular highlights become embedded in the surface normals, preserving the specific highlight shapes seen in training even when relighted with new environment maps. In contrast, our ReCap with dual lighting produce accurate highlight shapes with better surface normal.
\begin{figure}[h]
    \centering
    \includegraphics[width=1.0\linewidth]{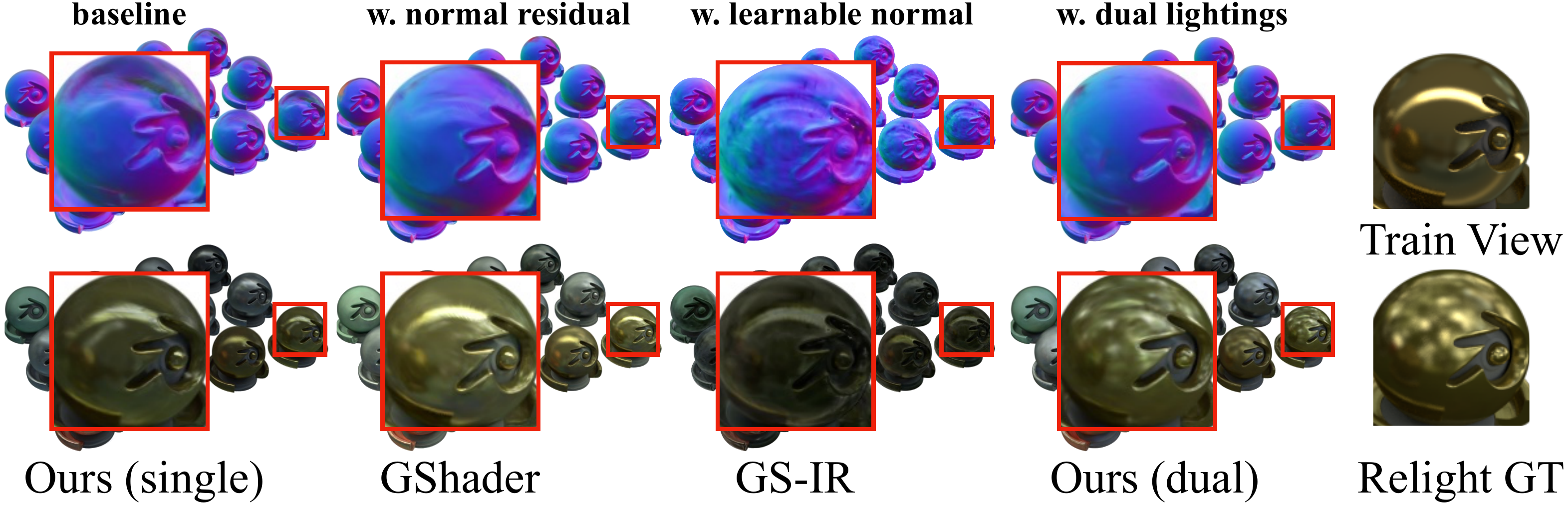}
    \vspace{-1.5em}
    \caption{The comparison of estimated normal and corresponding relighting results. With single-environment captures, the highlights from the train view are falsely attributed to object property instead of lighting, passing down to relighting views. Cross-environment supervision provides more robust normal estimation and correct highlight shapes.}
    \label{fig:normal}
    \vspace{-1.5em}
\end{figure}
%UPDATED
\begin{table*}[!ht]   
    \centering
    \scriptsize
    \setlength\tabcolsep{4.4pt}
    \caption{\fboxsep0.75pt Relighting and NVS results. Methods optimized over single-env (\scalebox{0.75}{\ding{70}}) and dual-env (\scalebox{0.75}{\ding{71}}) inputs differ in the choice of training split B. ``GT scaling" refers to the use of ground truth albedo to re-scale learned albedo for relighting. Best and second best results are highlighted in \textbf{\colorbox{lighttop1}{bold}} and \textit{\colorbox{lighttop2}{italic}} respectively.}
    \vspace{-1em}
    \begin{tabularx}{0.99\linewidth}{X*{9}{c}>{\columncolor{columnhighlight}}c *{4}{c}} % Define number of columns
        \hline
        
        \hline

        \hline
       &\multirow{2}{*}{\vspace{-3.5mm}\makecell{\textbf{use GT} \\\textbf{scaling}}}&\multicolumn{2}{c}{\textit{Training Envs}}&\multicolumn{7}{c}{\textit{Unseen Relights}}&\multicolumn{2}{c}{\textit{Seen Relights}}&\multicolumn{2}{c}{\textit{NVS}} \\
        \cmidrule(lr){3-4}\cmidrule(lr){5-11}\cmidrule(lr){12-13}\cmidrule(lr){14-15}
        & &\textbf{split A}&\textbf{split B}&\textbf{night}&\textbf{snow}&\textbf{sunset}&\textbf{interior}&\textbf{fireplace}&\textbf{forest}&\textbf{Avg.}&\textbf{bridge}&\textbf{courtyard}&\textbf{bridge}* &\textbf{courtyard*}
        \\
	\hline

	\hline
	\multicolumn{14}{c}{PSNR$\uparrow$}\\
	\hline
	3DGS-DR\textsuperscript{\ding{70}}\cite{ye20243d}&-&bridge&bridge&20.18&18.99&23.19&21.33&18.76&\cellcolor{lighttop2}\textit{24.90}&21.22&\cellcolor{lighttop1}\textbf{33.18}&-&\cellcolor{lighttop1}\textbf{35.59}&-
	\\
	GS-IR\textsuperscript{\ding{70}}\cite{liang2024gs}&-&bridge&bridge&25.56&\cellcolor{lighttop2}\textit{22.44}&21.22&19.46&25.64&21.10&22.57&22.01&-&31.22&-
	\\
	R3DG\textsuperscript{\ding{70}}\cite{R3DG2023}&-&bridge&bridge&23.71&22.09&22.84&20.09&23.08&22.19&22.33&22.32&-&30.61&-
	\\
	GShader\textsuperscript{\ding{70}}\cite{jiang2024gaussianshader}&-&bridge&bridge&21.48&17.61&18.93&21.89&20.18&22.70&20.47&26.03&-&33.05&-
	\\
	ReCap\textsuperscript{\ding{70}}~(ours)&-&bridge&bridge&25.31&21.76&23.20&21.66&24.52&23.15&23.27&25.30&-&\cellcolor{lighttop2}\textit{33.40}&-
	\\
	\hline
	3DGS-DR\textsuperscript{\ding{71}}\cite{ye20243d}&-&bridge&courtyard&20.53&20.25&\cellcolor{lighttop2}\textit{24.56}&21.71&19.41&24.25&21.78&24.89&23.92&24.84&23.57
	\\
	GS-IR\textsuperscript{\ding{71}}\cite{liang2024gs}&-&bridge&courtyard&26.22&22.03&20.85&19.18&\cellcolor{lighttop2}\textit{25.90}&20.52&22.45&20.78&20.96&24.92&23.16
	\\
	R3DG\textsuperscript{\ding{71}}\cite{R3DG2023}&-&bridge&courtyard&23.91&21.78&22.55&19.86&23.38&21.74&22.21&20.98&20.84&25.26&22.68
	\\
	GShader\textsuperscript{\ding{71}}\cite{jiang2024gaussianshader}&-&bridge&courtyard&22.73&18.20&19.96&22.11&20.97&23.02&21.17&23.34&21.75&24.96&22.84
	\\
	\hline
	TensoIR\textsuperscript{\ding{71}}\cite{jin2023tensoir}&\cmark&bridge&courtyard&\cellcolor{lighttop2}\textit{27.22}&22.15&24.31&23.12&25.35&24.80&\cellcolor{lighttop2}\textit{24.49}&24.50&23.82&29.13&27.46
	\\
	TensoIR\textsuperscript{\ding{71}}\textsubscript{no scale}\cite{jin2023tensoir}&-&bridge&courtyard&26.24&20.11&22.75&21.79&24.45&23.29&23.11&23.50&22.47&29.13&27.46
	\\
	GShader\textsuperscript{\ding{71}}\cite{jiang2024gaussianshader}~+~ours&-&bridge&courtyard&23.52&17.69&19.95&\cellcolor{lighttop2}\textit{23.22}&21.75&24.50&21.77&24.45&\cellcolor{lighttop2}\textit{24.32}&31.13&\cellcolor{lighttop2}\textit{29.30}
	\\
	ReCap\textsuperscript{\ding{71}}~(ours)&-&bridge&courtyard&\cellcolor{lighttop1}\textbf{27.62}&\cellcolor{lighttop1}\textbf{24.64}&\cellcolor{lighttop1}\textbf{26.33}&\cellcolor{lighttop1}\textbf{24.40}&\cellcolor{lighttop1}\textbf{26.52}&\cellcolor{lighttop1}\textbf{25.38}&\cellcolor{lighttop1}\textbf{25.82}&\cellcolor{lighttop2}\textit{26.95}&\cellcolor{lighttop1}\textbf{25.52}&32.23&\cellcolor{lighttop1}\textbf{30.76}
	\\
	\hline

	\hline
	\multicolumn{14}{c}{SSIM$\uparrow$}\\
	\hline
	3DGS-DR\textsuperscript{\ding{70}}\cite{ye20243d}&-&bridge&bridge&0.882&0.894&0.925&0.907&0.843&0.906&0.893&\cellcolor{lighttop1}\textbf{0.971}&-&\cellcolor{lighttop1}\textbf{0.978}&-
	\\
	GS-IR\textsuperscript{\ding{70}}\cite{liang2024gs}&-&bridge&bridge&0.902&0.904&0.890&0.861&0.884&0.867&0.885&0.898&-&0.952&-
	\\
	R3DG\textsuperscript{\ding{70}}\cite{R3DG2023}&-&bridge&bridge&0.889&0.913&0.918&0.866&0.866&0.877&0.888&0.879&-&0.959&-
	\\
	GShader\textsuperscript{\ding{70}}\cite{jiang2024gaussianshader}&-&bridge&bridge&0.889&0.887&0.896&0.905&0.854&0.904&0.889&0.949&-&0.968&-
	\\
	ReCap\textsuperscript{\ding{70}}~(ours)&-&bridge&bridge&0.911&\cellcolor{lighttop2}\textit{0.918}&0.926&0.907&0.883&0.906&\cellcolor{lighttop2}\textit{0.909}&0.946&-&\cellcolor{lighttop2}\textit{0.970}&-
	\\
	\hline
	3DGS-DR\textsuperscript{\ding{71}}\cite{ye20243d}&-&bridge&courtyard&0.885&0.902&\cellcolor{lighttop2}\textit{0.934}&0.907&0.846&0.903&0.896&0.930&0.927&0.926&0.924
	\\
	GS-IR\textsuperscript{\ding{71}}\cite{liang2024gs}&-&bridge&courtyard&0.906&0.893&0.887&0.858&0.885&0.862&0.882&0.881&0.879&0.915&0.905
	\\
	R3DG\textsuperscript{\ding{71}}\cite{R3DG2023}&-&bridge&courtyard&0.896&0.912&0.922&0.866&0.870&0.873&0.890&0.859&0.888&0.932&0.919
	\\
	GShader\textsuperscript{\ding{71}}\cite{jiang2024gaussianshader}&-&bridge&courtyard&0.904&0.891&0.908&0.906&0.862&0.905&0.896&0.925&0.913&0.928&0.922
	\\
	\hline
	TensoIR\textsuperscript{\ding{71}}\cite{jin2023tensoir}&\cmark&bridge&courtyard&0.908&0.861&0.891&0.870&0.886&0.888&0.884&0.893&0.883&0.962&\cellcolor{lighttop2}\textit{0.957}
	\\
	TensoIR\textsuperscript{\ding{71}}\textsubscript{no scale}\cite{jin2023tensoir}&-&bridge&courtyard&0.910&0.861&0.893&0.871&\cellcolor{lighttop2}\textit{0.889}&0.889&0.885&0.895&0.884&0.962&0.957
	\\
	GShader\textsuperscript{\ding{71}}\cite{jiang2024gaussianshader}~+~ours&-&bridge&courtyard&\cellcolor{lighttop2}\textit{0.915}&0.893&0.915&\cellcolor{lighttop2}\textit{0.919}&0.874&\cellcolor{lighttop2}\textit{0.922}&0.906&0.937&\cellcolor{lighttop2}\textit{0.938}&0.959&0.956
	\\
	ReCap\textsuperscript{\ding{71}}~(ours)&-&bridge&courtyard&\cellcolor{lighttop1}\textbf{0.935}&\cellcolor{lighttop1}\textbf{0.938}&\cellcolor{lighttop1}\textbf{0.951}&\cellcolor{lighttop1}\textbf{0.929}&\cellcolor{lighttop1}\textbf{0.903}&\cellcolor{lighttop1}\textbf{0.926}&\cellcolor{lighttop1}\textbf{0.930}&\cellcolor{lighttop2}\textit{0.951}&\cellcolor{lighttop1}\textbf{0.943}&0.966&\cellcolor{lighttop1}\textbf{0.963}
	\\
	\hline

	\hline
	\multicolumn{14}{c}{LPIPS$\downarrow$}\\
	\hline
	3DGS-DR\textsuperscript{\ding{70}}\cite{ye20243d}&-&bridge&bridge&0.081&0.091&0.073&\cellcolor{lighttop2}\textit{0.085}&0.108&0.083&0.087&\cellcolor{lighttop1}\textbf{0.041}&-&\cellcolor{lighttop1}\textbf{0.034}&-
	\\
	GS-IR\textsuperscript{\ding{70}}\cite{liang2024gs}&-&bridge&bridge&0.099&0.094&0.101&0.119&0.108&0.109&0.105&0.099&-&0.065&-
	\\
	R3DG\textsuperscript{\ding{70}}\cite{R3DG2023}&-&bridge&bridge&0.103&0.086&0.082&0.124&0.114&0.104&0.102&0.113&-&0.053&-
	\\
	GShader\textsuperscript{\ding{70}}\cite{jiang2024gaussianshader}&-&bridge&bridge&0.091&0.116&0.098&0.094&0.113&0.091&0.100&0.062&-&0.044&-
	\\
	ReCap\textsuperscript{\ding{70}}~(ours)&-&bridge&bridge&0.077&\cellcolor{lighttop2}\textit{0.080}&\cellcolor{lighttop2}\textit{0.073}&0.089&\cellcolor{lighttop2}\textit{0.092}&0.084&\cellcolor{lighttop2}\textit{0.083}&0.061&-&\cellcolor{lighttop2}\textit{0.042}&-
	\\
	\hline
	3DGS-DR\textsuperscript{\ding{71}}\cite{ye20243d}&-&bridge&courtyard&0.084&0.096&0.078&0.096&0.115&0.094&0.094&0.078&0.080&0.082&0.085
	\\
	GS-IR\textsuperscript{\ding{71}}\cite{liang2024gs}&-&bridge&courtyard&0.101&0.107&0.107&0.127&0.112&0.116&0.112&0.115&0.119&0.099&0.104
	\\
	R3DG\textsuperscript{\ding{71}}\cite{R3DG2023}&-&bridge&courtyard&0.101&0.092&0.083&0.129&0.117&0.110&0.105&0.127&0.106&0.078&0.089
	\\
	GShader\textsuperscript{\ding{71}}\cite{jiang2024gaussianshader}&-&bridge&courtyard&0.087&0.115&0.095&0.101&0.114&0.097&0.101&0.086&0.094&0.080&0.086
	\\
	\hline
	TensoIR\textsuperscript{\ding{71}}\cite{jin2023tensoir}&\cmark&bridge&courtyard&0.131&0.155&0.134&0.138&0.138&0.127&0.137&0.128&0.133&0.093&0.098
	\\
	TensoIR\textsuperscript{\ding{71}}\textsubscript{no scale}\cite{jin2023tensoir}&-&bridge&courtyard&0.133&0.159&0.136&0.141&0.138&0.131&0.140&0.130&0.137&0.093&0.098
	\\
	GShader\textsuperscript{\ding{71}}\cite{jiang2024gaussianshader}~+~ours&-&bridge&courtyard&\cellcolor{lighttop2}\textit{0.074}&0.113&0.084&0.089&0.102&\cellcolor{lighttop2}\textit{0.080}&0.090&0.073&\cellcolor{lighttop2}\textit{0.073}&0.055&\cellcolor{lighttop2}\textit{0.059}
	\\
	ReCap\textsuperscript{\ding{71}}~(ours)&-&bridge&courtyard&\cellcolor{lighttop1}\textbf{0.059}&\cellcolor{lighttop1}\textbf{0.069}&\cellcolor{lighttop1}\textbf{0.057}&\cellcolor{lighttop1}\textbf{0.077}&\cellcolor{lighttop1}\textbf{0.075}&\cellcolor{lighttop1}\textbf{0.071}&\cellcolor{lighttop1}\textbf{0.068}&\cellcolor{lighttop2}\textit{0.058}&\cellcolor{lighttop1}\textbf{0.064}&0.047&\cellcolor{lighttop1}\textbf{0.051}
	\\
        \hline

        \hline

        \hline
    \end{tabularx}
    \label{tab:scene_wise}
    \vspace{-1.5em}
\end{table*}
\section{Experiments}
\label{sec:exp}
\subsection{Implementation Details}
\noindent{\textbf{Datasets.}}
We construct a more comprehensive \textit{RelightObj} dataset by relighting 8 general objects from NeRF Synthetic Dataset~\cite{mildenhall2020nerf} and 5 shiny objects from the Shiny Blender Dataset~\cite{verbin2022ref} under 8 different HDR maps. Each scene includes 200 training views and 200 test views, with identical camera poses across scenes. 
To eliminate biases from training view poses, the training views are divided into two splits, A and B, each containing 100 views. In the single-environment (single-env) setup, both splits come from the same environment, whereas in the dual-environment (dual-env) setup, split B is taken from a different environment. Relighting performance is judged on unseen environments. 
\vspace{-1em}

\noindent{\textbf{Baseline and Metrics.}} We compare with the following baselines: 
(a) \textbf{GShader}~\cite{jiang2024gaussianshader}:  a Gaussian rendering method utilizing a customized shading function with a residual color term;
(b) \textbf{GS-IR}~\cite{liang2024gs}: an inverse rendering approach using view-space shading with a microfacet BRDF model and a baked occlusion map;
(c) \textbf{3DGS-DR}~\cite{ye20243d}: a Gasussian rendering method targeting reflective surfaces with a customized shading function and normal propagation;
(d) \textbf{R3DG}~\cite{R3DG2023}: a Gaussian relighting technique that incorporates ray tracing and a microfacet BRDF model; and 
(e) \textbf{TensoIR}~\cite{jin2023tensoir}: top-performing NeRF-based method in novel view synthesis and relighting. Following these works, rendering and relighting quality are evaluated using PSNR, SSIM~\cite{wang2004image}, and LPIPS~\cite{zhang2018unreasonable}.

\noindent{\textbf{Training and Testing.}}
All experiments are conducted on an Nvidia RTX 3090 graphics card. Our models are optimized using Adam for 30,000 iterations. All other methods are retrained on our dataset using their official repositories and recommended settings, with results reported from the retrained models. As GShader~\cite{jiang2024gaussianshader} and 3DGS-DR~\cite{ye20243d} lack official relighting code, we pre-processed HDR maps following their design choices (details in the supplementary). For TensoIR~\cite{jin2023tensoir}, which uses per-object hyperparameters, we tested all provided settings and report the best results.
\begin{figure*}
    \centering
    \includegraphics[width=0.89\textwidth,]{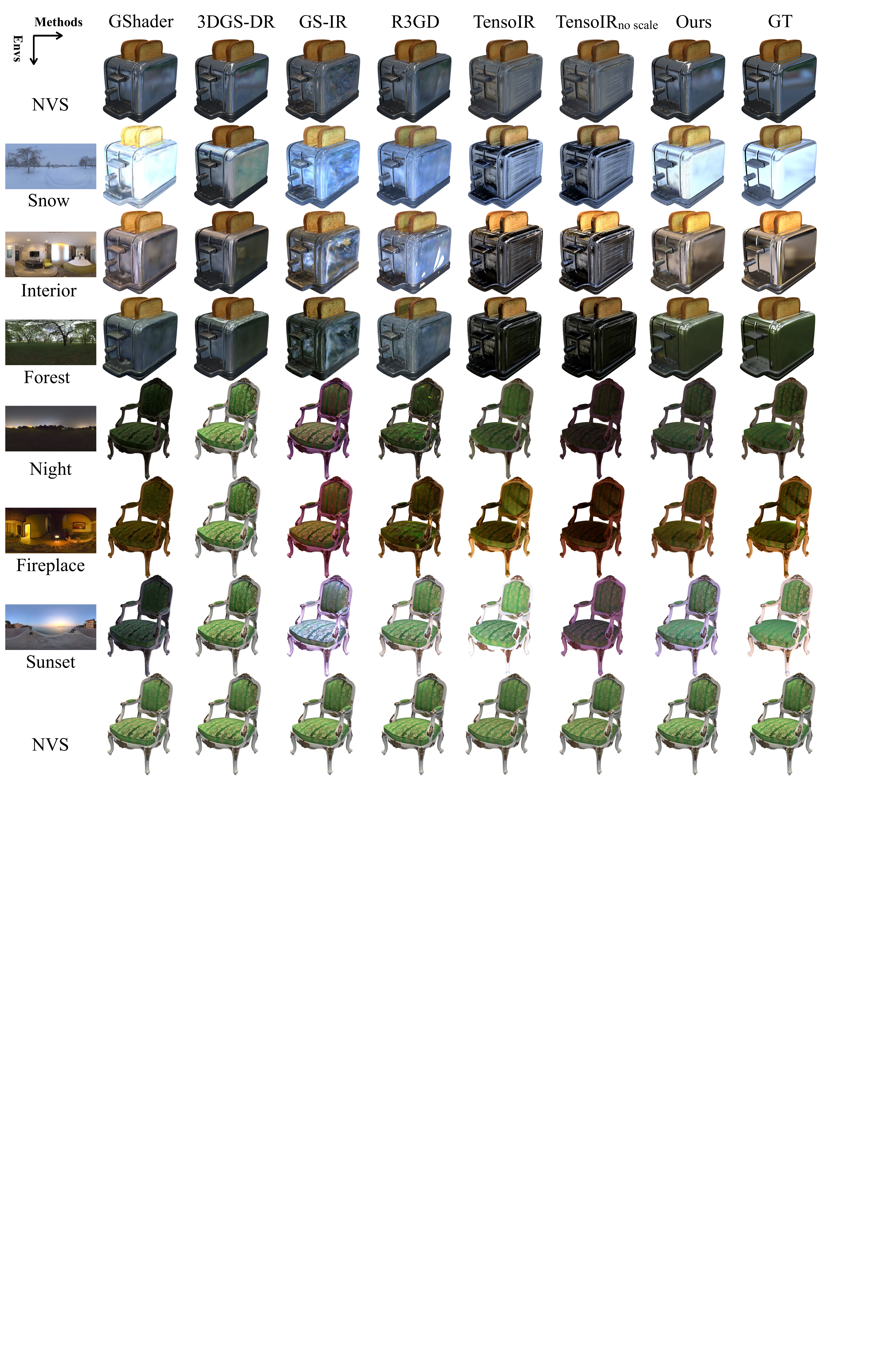}
    \vspace{-1em}
    \caption{Qualitative comparison of NVS and relighting results. Upper: \textit{toaster}, lower: \textit{chair}. TensoIR produces accurate relighting with GT albedo scaling for \textit{chair}, but suffers from tone-shift without it. For our method, specular surfaces in \textit{toaster} provide more information for lighting reconstruction, leading to better albedo-lighting decoupling hence more truthful relighting results than mostly diffuse \textit{chair}. Refer to the supplementary for more visuals.}
    \label{fig:nvs_vs_relight}
    \vspace{-1.5em}
\end{figure*}
\subsection{Comparison with Previous Methods}
\noindent{\textbf{Single-env Setup.}} We compare single-env performance in the first group of~\cref{tab:scene_wise}. The substantial difference between NVS using the learned environment map and relighting using the ground truth HDR map (bridge* vs. bridge) highlights the albedo-lighting ambiguity: the learned lighting representation diverges from ground truth but still works effectively with the learned materials for NVS. While 3DGS-DR\textsuperscript{\ding{70}}~\cite{ye20243d} has the smallest gap, its ability to generalize to unseen relighting conditions remains limited. As shown in the second column of~\cref{fig:nvs_vs_relight}, this is because it tends to reproduce training views, preserving the overall tone and adapting only the specular reflections to new lighting.

\noindent{\textbf{Dual-env Setup.}} The second group of~\cref{tab:scene_wise} presents results from directly optimizing existing Gaussian models over dual-env inputs while maintaining a single learned lighting representation. This leads to mixed relighting performance and significant NVS artifacts since the learned environment is no longer well-defined. Therefore, we continue our analysis with their single-env results for the remainder of the discussion. In the third group, we compare against TensoIR~\cite{jin2023tensoir} with their multi-light setup and further augment GShader~\cite{jiang2024gaussianshader} with the proposed multi-light querying. Since TensoIR~\cite{jin2023tensoir} uses GT albedo scaling in their official code, we additionally report their scale-free performance for clearer comparison.  

\noindent{\textbf{Qualitative Comparison.}} \cref{fig:nvs_vs_relight} compares NVS and relighting performance across single-env Gaussian relighting methods, two dual-env TensoIR~\cite{jin2023tensoir} variants, and dual-env ReCap. Our ReCap achieves noticeable perceptual improvements over methods that do not rely on GT for relighting. TensoIR~\cite{jin2023tensoir} produces accurate relighting with GT albedo scaling, but suffers from obvious tone-shift without it (e.g., chair) and struggles with metallic objects (e.g., toaster) due to their dielectric assumption.

\noindent{\textbf{Reconstructed Environments.}}
While specular objects pose significant challenges for NVS due to their complex view-dependent appearance\cite{jiang2024gaussianshader,ye20243d}, they offer distinct advantages for relighting tasks. Specular surfaces act as natural light probes, providing abundant high-frequency, view-dependent details that facilitate more accurate environment map reconstruction.  
In \cref{fig:env_comp}, we display the reconstructed environment maps for objects with varying level of specularity. For predominantly diffuse objects, disentangling lighting from intrinsic properties is challenging due to limited lighting cues. Notice how our learned light map for \textit{chair} retains object color, highlighting this challenge. 

\begin{figure}[t]
    \centering
    \includegraphics[width=1.0\linewidth,]{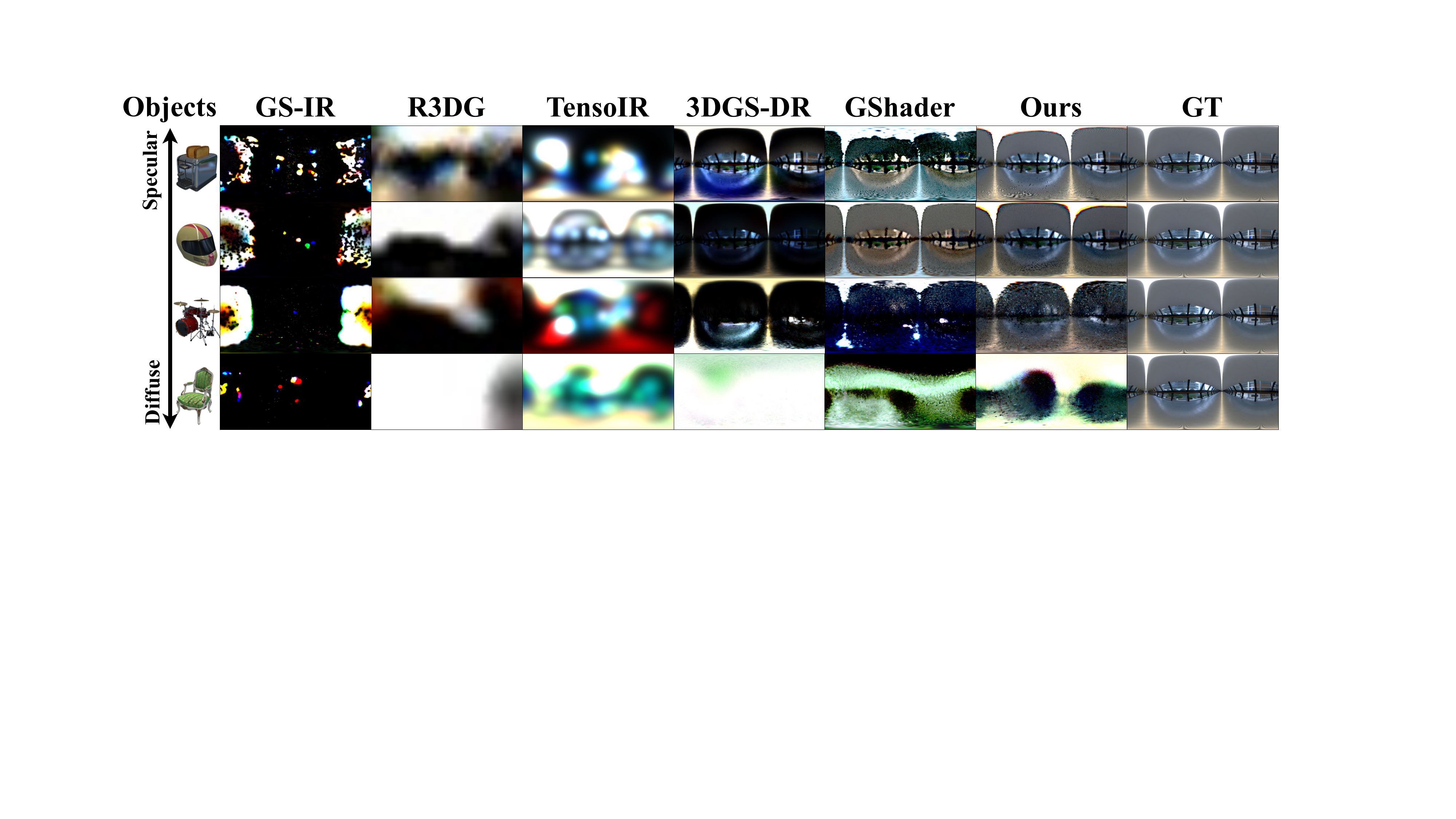}
    \vspace{-1.5em}
    \caption{Learned environment maps for different objects. As objects become more diffuse, reconstructions are less accurate.}
    \vspace{-1.5em}
    \label{fig:env_comp}
\end{figure}

\subsection{Ablation Study}
\label{sec:ablate}
\noindent{\textbf{Shading Function.}}
To demonstrate the effectiveness of our proposed shading function in~\cref{eq:ownshading}, we compare the average relighting performance across unseen scenes in ~\cref{tab:ablate_shading}. The new shading function yields the most substantial improvement by offering a more flexible material representation, with the two additional regularization terms also contributing positively to the final results. Additional visual examples are provided in the supplementary material.
\begin{table}[h]
    \footnotesize
    \centering
    \setlength\tabcolsep{7pt}
    \vspace{-1em}
    \caption{Ablation of shading function and regularization terms. $\mathcal{L}_{sat}$: specular saturation loss. $\mathcal{L}_{ec}$: energy conservation loss.}
    \vspace{-1em}
    \resizebox{.85\linewidth}{!}{
    \begin{tabular}{cccccc}
         \toprule
         Proposed Shading&$\mathcal{L}_{sat}$&$\mathcal{L}_{ec}$&PSNR&SSIM&LPIPS  \\
         \hline
         -&-&-&24.62&0.923&0.076\\
         \cmark&-&-&25.29&0.929&0.070\\
         \cmark& \cmark &-&25.38&0.929&0.069\\
         \rowcolor{rowhighlight!50}\cmark& \cmark &\cmark&\textbf{25.58}&\textbf{0.930}&\textbf{0.069}\\
         \bottomrule
    \end{tabular}}
    \label{tab:ablate_shading}
    \vspace{-1em}
\end{table}

\noindent{\textbf{Extra training environments.}}
So far, we have focused on dual-env setups. Here, we examine how extra photometric supervision from additional training environments affects the relighting. Due to the constraint of 200 training views per scene, we allocate 100 identical views per environment as we include more environments for training. To align with existing results, we also compare the use of identical views versus extra views. 
Results in~\cref{tab:ablate_envs} reveal three key findings: 
(i) Adding identical views from an extra environment is better than adding extra unique views within the same environment (row 2 vs.~3); (ii) Although identical views across environments theoretically provides stronger decoupling, the benefit of extra unique views is more pronounced in practice (row 3 vs.~4). It is likely that extra views already provide sufficient decoupling, especially with perfect pose calibration in synthetic datasets; and (iii) Increasing the number of environments consistently enhances relighting performance, with no observed plateau up to five.

\begin{table}[t]
    \footnotesize
    \centering
    \caption{Ablation on the number of environments used in training. Since up to $5$ environments are used, relighting results are reported on the remaining 3 unseen scenes: sunset, fireplace, night.}
    \vspace{-1em}
    \resizebox{0.91\linewidth}{!}{\begin{tabular}{ccccccc}
         \toprule
         \# Envs&\# Views&\# Unique Views& Cam Poses&PSNR&SSIM&LPIPS  \\
         \hline
         1& 100 &100 &identical&23.98&0.906&0.084\\
         \rowcolor{rowhighlight!50}1& 200 &200 &extra&24.07&0.908&0.081\\
         2& 200 &100 &identical&26.14&0.927&0.066\\
         \rowcolor{rowhighlight!50}2&200&200 &extra&26.25&0.928&0.065\\
         3&300&100 &identical&26.35&0.930&0.064\\
         4&400&100 &identical&26.68&0.931&0.062\\
         \rowcolor{rowhighlight!50}5&500&100 &identical&27.36&0.936&0.060\\
         \bottomrule
    \end{tabular}}
    \label{tab:ablate_envs}
    \vspace{-2em}
\end{table}

\noindent{\textbf{Application on Real Captures.}} ReCap requires multiple sets of camera poses to align to a common coordinate. In~\cref{fig:practicality}, we demonstrate its practicality with two real-life examples from StanfordORB~\cite{kuang2023stanford}, which captures the same object in different environments and provides COLMAP-estimated poses for each environment. They also provide sparse cross-environment alignment pairs computed by SuperGlue~\cite{sarlin2020superglue}. We use these pairs to estimate the view transform matrix and account for scale ambiguity. While exact calibration is non-trivial, sufficiently accurate pose calibration proves to be feasible for the proposed application.
\begin{figure}[h]
    \centering
    \vspace{-0.5em}
    \includegraphics[width=1.0\linewidth]{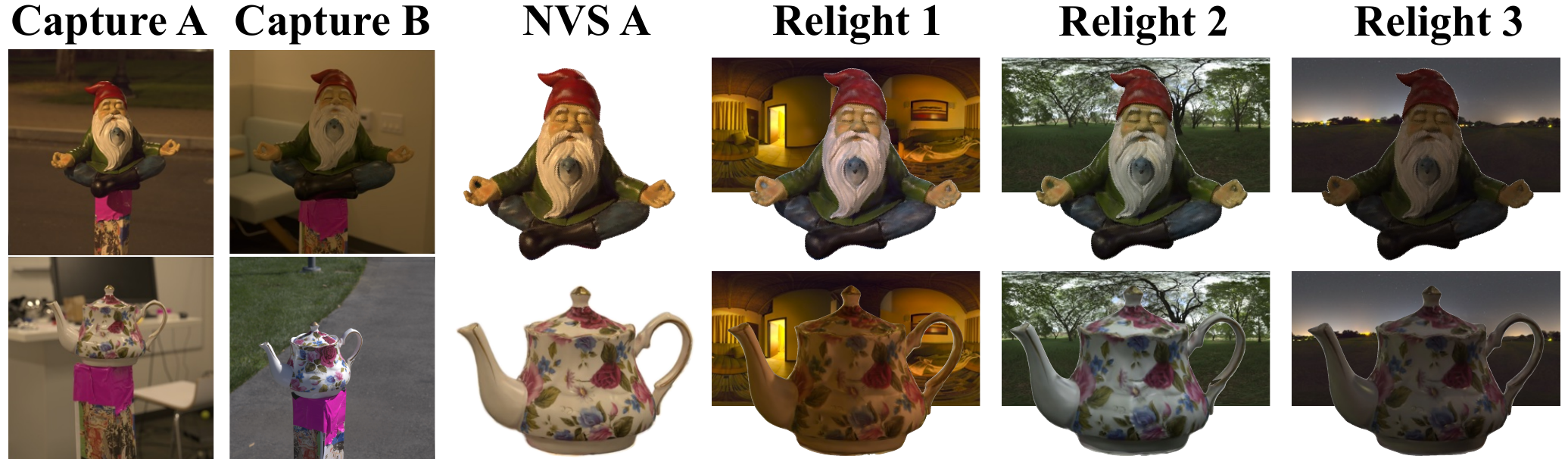}
    \caption{Relighting and NVS examples on StanfordORB: \textit{gnome}, \textit{teapot}.}
    \vspace{-1em}
    \label{fig:practicality}
\end{figure}
\subsection{Limitation and discussion}
While we achieve compelling relighting results, indirect illumination and subsurface scattering effects are not considered. Over-exposed regions also present inherent challenges for opacity estimation in all Gaussian-based relighting methods, where clipped highlights are incorrectly interpreted as transparent against white splatting backgrounds. For easier pose calibration in practice, semi-controlled capturing, such as rotating the object within the same environment at a known angle, may be considered.
\section{Conclusion}
In this paper, we proposed ReCap, leveraging internal photometric consistency to address the albedo-lighting ambiguity limiting existing Gaussian relighting methods. With cross-environment captures, we explicitly model light-dependent appearance with independent learnable lighting representations that share a common set of material attributes. Combined with an optimization-friendly shading function and physically appropriate post-processing during training, ReCap demonstrates realistic relighting quality with truthful tones across diverse scenes.

\vfill
\noindent{\textbf{Acknowledgment.}}
This research was supported by the Alexander von Humboldt Foundation.

\clearpage
{
    \small
    \bibliographystyle{ieeenat_fullname}
    \bibliography{main}

\begin{thebibliography}{51}
\providecommand{\natexlab}[1]{#1}
\providecommand{\url}[1]{\texttt{#1}}
\expandafter\ifx\csname urlstyle\endcsname\relax
  \providecommand{\doi}[1]{doi: #1}\else
  \providecommand{\doi}{doi: \begingroup \urlstyle{rm}\Url}\fi

\bibitem[Barron and Malik(2012)]{barron2012shape}
Jonathan~T Barron and Jitendra Malik.
\newblock Shape, albedo, and illumination from a single image of an unknown object.
\newblock In \emph{CVPR}, 2012.

\bibitem[Barron et~al.(2021)Barron, Mildenhall, Tancik, Hedman, Martin-Brualla, and Srinivasan]{barron2021mip}
Jonathan~T Barron, Ben Mildenhall, Matthew Tancik, Peter Hedman, Ricardo Martin-Brualla, and Pratul~P Srinivasan.
\newblock Mip-nerf: A multiscale representation for anti-aliasing neural radiance fields.
\newblock In \emph{ICCV}, 2021.

\bibitem[Barron et~al.(2022)Barron, Mildenhall, Verbin, Srinivasan, and Hedman]{barron2022mip}
Jonathan~T Barron, Ben Mildenhall, Dor Verbin, Pratul~P Srinivasan, and Peter Hedman.
\newblock Mip-nerf 360: Unbounded anti-aliased neural radiance fields.
\newblock In \emph{CVPR}, 2022.

\bibitem[Barron et~al.(2023)Barron, Mildenhall, Verbin, Srinivasan, and Hedman]{barron2023zip}
Jonathan~T Barron, Ben Mildenhall, Dor Verbin, Pratul~P Srinivasan, and Peter Hedman.
\newblock Zip-nerf: Anti-aliased grid-based neural radiance fields.
\newblock In \emph{ICCV}, 2023.

\bibitem[Barrow et~al.(1978)Barrow, Tenenbaum, Hanson, and Riseman]{barrow1978recovering}
Harry Barrow, J Tenenbaum, A Hanson, and E Riseman.
\newblock Recovering intrinsic scene characteristics.
\newblock \emph{Computer Vision Systems}, 2\penalty0 (3-26):\penalty0 2, 1978.

\bibitem[Bi et~al.(2020)Bi, Xu, Sunkavalli, Ha{\v{s}}an, Hold-Geoffroy, Kriegman, and Ramamoorthi]{bi2020deep}
Sai Bi, Zexiang Xu, Kalyan Sunkavalli, Milo{\v{s}} Ha{\v{s}}an, Yannick Hold-Geoffroy, David Kriegman, and Ravi Ramamoorthi.
\newblock Deep reflectance volumes: Relightable reconstructions from multi-view photometric images.
\newblock In \emph{ECCV}, 2020.

\bibitem[Boss et~al.(2021)Boss, Braun, Jampani, Barron, Liu, and Lensch]{boss2021nerd}
Mark Boss, Raphael Braun, Varun Jampani, Jonathan~T Barron, Ce Liu, and Hendrik Lensch.
\newblock Nerd: Neural reflectance decomposition from image collections.
\newblock In \emph{ICCV}, pages 12684--12694, 2021.

\bibitem[Burley and Studios(2012)]{burley2012physically}
Brent Burley and Walt Disney~Animation Studios.
\newblock Physically-based shading at disney.
\newblock In \emph{ACM SIGGRAPH}, pages 1--7. vol. 2012, 2012.

\bibitem[Cook and Torrance(1981)]{cook1981reflectance}
Robert~L Cook and Kenneth~E Torrance.
\newblock A reflectance model for computer graphics.
\newblock \emph{ACM SIGGRAPH}, 15\penalty0 (3):\penalty0 307--316, 1981.

\bibitem[Debevec(1998)]{debevec1998rendering}
Paul Debevec.
\newblock Rendering with natural light.
\newblock In \emph{ACM SIGGRAPH Electronic art and animation catalog}, page 166. 1998.

\bibitem[Debevec(2012)]{debevec2012light}
Paul Debevec.
\newblock The light stages and their applications to photoreal digital actors.
\newblock \emph{ACM SIGGRAPH Asia}, 2\penalty0 (4):\penalty0 1--6, 2012.

\bibitem[Debevec et~al.(2000)Debevec, Hawkins, Tchou, Duiker, Sarokin, and Sagar]{debevec2000acquiring}
Paul Debevec, Tim Hawkins, Chris Tchou, Haarm-Pieter Duiker, Westley Sarokin, and Mark Sagar.
\newblock Acquiring the reflectance field of a human face.
\newblock In \emph{PACMCGIT}, pages 145--156, 2000.

\bibitem[Feng et~al.(2022)Feng, Bolkart, Tesch, Black, and Abrevaya]{feng2022towards}
Haiwen Feng, Timo Bolkart, Joachim Tesch, Michael~J Black, and Victoria Abrevaya.
\newblock Towards racially unbiased skin tone estimation via scene disambiguation.
\newblock In \emph{ECCV}, 2022.

\bibitem[Fridovich-Keil et~al.(2022)Fridovich-Keil, Yu, Tancik, Chen, Recht, and Kanazawa]{fridovich2022plenoxels}
Sara Fridovich-Keil, Alex Yu, Matthew Tancik, Qinhong Chen, Benjamin Recht, and Angjoo Kanazawa.
\newblock Plenoxels: Radiance fields without neural networks.
\newblock In \emph{CVPR}, 2022.

\bibitem[Gao et~al.(2024)Gao, Gu, Lin, Zhu, Cao, Zhang, and Yao]{R3DG2023}
Jian Gao, Chun Gu, Youtian Lin, Hao Zhu, Xun Cao, Li Zhang, and Yao Yao.
\newblock Relightable 3d gaussians: Realistic point cloud relighting with brdf decomposition and ray tracing.
\newblock In \emph{ECCV}, 2024.

\bibitem[Garon et~al.(2019)Garon, Sunkavalli, Hadap, Carr, and Lalonde]{garon2019fast}
Mathieu Garon, Kalyan Sunkavalli, Sunil Hadap, Nathan Carr, and Jean-Fran{\c{c}}ois Lalonde.
\newblock Fast spatially-varying indoor lighting estimation.
\newblock In \emph{CVPR}, 2019.

\bibitem[Gu{\'e}don and Lepetit(2024)]{guedon2024sugar}
Antoine Gu{\'e}don and Vincent Lepetit.
\newblock Sugar: Surface-aligned gaussian splatting for efficient 3d mesh reconstruction and high-quality mesh rendering.
\newblock In \emph{CVPR}, 2024.

\bibitem[Jiang et~al.(2024)Jiang, Tu, Liu, Gao, Long, Wang, and Ma]{jiang2024gaussianshader}
Yingwenqi Jiang, Jiadong Tu, Yuan Liu, Xifeng Gao, Xiaoxiao Long, Wenping Wang, and Yuexin Ma.
\newblock Gaussianshader: 3d gaussian splatting with shading functions for reflective surfaces.
\newblock In \emph{CVPR}, 2024.

\bibitem[Jin et~al.(2023)Jin, Liu, Xu, Zhang, Han, Bi, Zhou, Xu, and Su]{jin2023tensoir}
Haian Jin, Isabella Liu, Peijia Xu, Xiaoshuai Zhang, Songfang Han, Sai Bi, Xiaowei Zhou, Zexiang Xu, and Hao Su.
\newblock Tensoir: Tensorial inverse rendering.
\newblock In \emph{CVPR}, 2023.

\bibitem[Kang et~al.(2019)Kang, Xie, He, Yi, Gu, Chen, Zhou, and Wu]{kang2019learning}
Kaizhang Kang, Cihui Xie, Chengan He, Mingqi Yi, Minyi Gu, Zimin Chen, Kun Zhou, and Hongzhi Wu.
\newblock Learning efficient illumination multiplexing for joint capture of reflectance and shape.
\newblock \emph{ACM TOG}, 38\penalty0 (6):\penalty0 165--1, 2019.

\bibitem[Karis and Games(2013)]{karis2013real}
Brian Karis and Epic Games.
\newblock Real shading in unreal engine 4.
\newblock \emph{Proc. Physically Based Shading Theory Practice}, 4\penalty0 (3):\penalty0 1, 2013.

\bibitem[Kerbl et~al.(2023)Kerbl, Kopanas, Leimk{\"u}hler, and Drettakis]{kerbl20233d}
Bernhard Kerbl, Georgios Kopanas, Thomas Leimk{\"u}hler, and George Drettakis.
\newblock 3d gaussian splatting for real-time radiance field rendering.
\newblock \emph{ACM TOG}, 42\penalty0 (4):\penalty0 139--1, 2023.

\bibitem[Kopanas et~al.(2021)Kopanas, Philip, Leimk{\"u}hler, and Drettakis]{kopanas2021point}
Georgios Kopanas, Julien Philip, Thomas Leimk{\"u}hler, and George Drettakis.
\newblock Point-based neural rendering with per-view optimization.
\newblock In \emph{CGF}, pages 29--43. Wiley Online Library, 2021.

\bibitem[Kuang et~al.(2023)Kuang, Zhang, Yu, Agarwala, Wu, Wu, et~al.]{kuang2023stanford}
Zhengfei Kuang, Yunzhi Zhang, Hong-Xing Yu, Samir Agarwala, Elliott Wu, Jiajun Wu, et~al.
\newblock Stanford-orb: a real-world 3d object inverse rendering benchmark.
\newblock \emph{NeurIPS}, 36:\penalty0 46938--46957, 2023.

\bibitem[Li et~al.(2020)Li, Zhou, Wu, Shi, Diao, and Tan]{li2020mvps}
Min Li, Zhenglong Zhou, Zhe Wu, Boxin Shi, Changyu Diao, and Ping Tan.
\newblock Multi-view photometric stereo: A robust solution and benchmark dataset for spatially varying isotropic materials.
\newblock \emph{IEEE TIP}, 29:\penalty0 4159--4173, 2020.

\bibitem[Liang et~al.(2024)Liang, Zhang, Feng, Shan, and Jia]{liang2024gs}
Zhihao Liang, Qi Zhang, Ying Feng, Ying Shan, and Kui Jia.
\newblock Gs-ir: 3d gaussian splatting for inverse rendering.
\newblock In \emph{CVPR}, 2024.

\bibitem[Liu et~al.(2023)Liu, Wang, Lin, Long, Wang, Liu, Komura, and Wang]{liu2023nero}
Yuan Liu, Peng Wang, Cheng Lin, Xiaoxiao Long, Jiepeng Wang, Lingjie Liu, Taku Komura, and Wenping Wang.
\newblock Nero: Neural geometry and brdf reconstruction of reflective objects from multiview images.
\newblock \emph{ACM TOG}, 42\penalty0 (4):\penalty0 1--22, 2023.

\bibitem[Mildenhall et~al.(2020)Mildenhall, Srinivasan, Tancik, Barron, Ramamoorthi, and Ng]{mildenhall2020nerf}
B Mildenhall, PP Srinivasan, M Tancik, JT Barron, R Ramamoorthi, and R Ng.
\newblock Nerf: Representing scenes as neural radiance fields for view synthesis.
\newblock In \emph{ECCV}, 2020.

\bibitem[M{\"u}ller et~al.(2022)M{\"u}ller, Evans, Schied, and Keller]{muller2022instant}
Thomas M{\"u}ller, Alex Evans, Christoph Schied, and Alexander Keller.
\newblock Instant neural graphics primitives with a multiresolution hash encoding.
\newblock \emph{ACM TOG}, 41\penalty0 (4):\penalty0 1--15, 2022.

\bibitem[Munkberg et~al.(2022)Munkberg, Hasselgren, Shen, Gao, Chen, Evans, M{\"u}ller, and Fidler]{munkberg2022extracting}
Jacob Munkberg, Jon Hasselgren, Tianchang Shen, Jun Gao, Wenzheng Chen, Alex Evans, Thomas M{\"u}ller, and Sanja Fidler.
\newblock Extracting triangular 3d models, materials, and lighting from images.
\newblock In \emph{CVPR}, 2022.

\bibitem[Nam et~al.(2018)Nam, Lee, Gutierrez, and Kim]{nam2018practical}
Giljoo Nam, Joo~Ho Lee, Diego Gutierrez, and Min~H Kim.
\newblock Practical svbrdf acquisition of 3d objects with unstructured flash photography.
\newblock \emph{ACM TOG}, 37\penalty0 (6):\penalty0 1--12, 2018.

\bibitem[Park et~al.(2020)Park, Holynski, and Seitz]{park2020seeing}
Jeong~Joon Park, Aleksander Holynski, and Steven~M Seitz.
\newblock Seeing the world in a bag of chips.
\newblock In \emph{Proceedings of the IEEE/CVF Conference on Computer Vision and Pattern Recognition}, pages 1417--1427, 2020.

\bibitem[Pintus et~al.(2019)Pintus, Dulecha, Ciortan, Gobbetti, and Giachetti]{pintus2019state}
Ruggero Pintus, Tinsae~Gebrechristos Dulecha, Irina Ciortan, Enrico Gobbetti, and Andrea Giachetti.
\newblock State-of-the-art in multi-light image collections for surface visualization and analysis.
\newblock In \emph{CGF}, pages 909--934. Wiley Online Library, 2019.

\bibitem[Ramamoorthi and Hanrahan(2001{\natexlab{a}})]{ramamoorthi2001efficient}
Ravi Ramamoorthi and Pat Hanrahan.
\newblock An efficient representation for irradiance environment maps.
\newblock In \emph{ACM SIGGRAPH}, 2001{\natexlab{a}}.

\bibitem[Ramamoorthi and Hanrahan(2001{\natexlab{b}})]{ramamoorthi2001signal}
Ravi Ramamoorthi and Pat Hanrahan.
\newblock A signal-processing framework for inverse rendering.
\newblock In \emph{ACM SIGGRAPH}, pages 117--128, 2001{\natexlab{b}}.

\bibitem[Ren et~al.(2015)Ren, Dong, Lin, Tong, and Guo]{ren2015image}
Peiran Ren, Yue Dong, Stephen Lin, Xin Tong, and Baining Guo.
\newblock Image based relighting using neural networks.
\newblock \emph{ACM TOG}, 34\penalty0 (4):\penalty0 1--12, 2015.

\bibitem[Ren et~al.(2023)Ren, Deng, Ma, Yan, and Yang]{ren2023improving}
Xingyu Ren, Jiankang Deng, Chao Ma, Yichao Yan, and Xiaokang Yang.
\newblock Improving fairness in facial albedo estimation via visual-textual cues.
\newblock In \emph{CVPR}, 2023.

\bibitem[Sarlin et~al.(2020)Sarlin, DeTone, Malisiewicz, and Rabinovich]{sarlin2020superglue}
Paul-Edouard Sarlin, Daniel DeTone, Tomasz Malisiewicz, and Andrew Rabinovich.
\newblock Superglue: Learning feature matching with graph neural networks.
\newblock In \emph{CVPR}, 2020.

\bibitem[Srinivasan et~al.(2021)Srinivasan, Deng, Zhang, Tancik, Mildenhall, and Barron]{srinivasan2021nerv}
Pratul~P Srinivasan, Boyang Deng, Xiuming Zhang, Matthew Tancik, Ben Mildenhall, and Jonathan~T Barron.
\newblock Nerv: Neural reflectance and visibility fields for relighting and view synthesis.
\newblock In \emph{CVPR}, 2021.

\bibitem[Verbin et~al.(2022)Verbin, Hedman, Mildenhall, Zickler, Barron, and Srinivasan]{verbin2022ref}
Dor Verbin, Peter Hedman, Ben Mildenhall, Todd Zickler, Jonathan~T Barron, and Pratul~P Srinivasan.
\newblock Ref-nerf: Structured view-dependent appearance for neural radiance fields.
\newblock In \emph{CVPR}, 2022.

\bibitem[Wang et~al.(2009)Wang, Ren, Gong, Snyder, and Guo]{wang2009all}
Jiaping Wang, Peiran Ren, Minmin Gong, John Snyder, and Baining Guo.
\newblock All-frequency rendering of dynamic, spatially-varying reflectance.
\newblock In \emph{ACM SIGGRAPH Asia}, 2009.

\bibitem[Wang et~al.(2004)Wang, Bovik, Sheikh, and Simoncelli]{wang2004image}
Zhou Wang, Alan~C Bovik, Hamid~R Sheikh, and Eero~P Simoncelli.
\newblock Image quality assessment: from error visibility to structural similarity.
\newblock \emph{IEEE TIP}, 13\penalty0 (4):\penalty0 600--612, 2004.

\bibitem[Xu et~al.(2022)Xu, Xu, Philip, Bi, Shu, Sunkavalli, and Neumann]{xu2022point}
Qiangeng Xu, Zexiang Xu, Julien Philip, Sai Bi, Zhixin Shu, Kalyan Sunkavalli, and Ulrich Neumann.
\newblock Point-nerf: Point-based neural radiance fields.
\newblock In \emph{CVPR}, 2022.

\bibitem[Xu et~al.(2018)Xu, Sunkavalli, Hadap, and Ramamoorthi]{xu2018deep}
Zexiang Xu, Kalyan Sunkavalli, Sunil Hadap, and Ravi Ramamoorthi.
\newblock Deep image-based relighting from optimal sparse samples.
\newblock \emph{ACM TOG}, 37\penalty0 (4):\penalty0 1--13, 2018.

\bibitem[Xu et~al.(2019)Xu, Bi, Sunkavalli, Hadap, Su, and Ramamoorthi]{xu2019deep}
Zexiang Xu, Sai Bi, Kalyan Sunkavalli, Sunil Hadap, Hao Su, and Ravi Ramamoorthi.
\newblock Deep view synthesis from sparse photometric images.
\newblock \emph{ACM TOG}, 38\penalty0 (4):\penalty0 1--13, 2019.

\bibitem[Ye et~al.(2024)Ye, Hou, and Zhou]{ye20243d}
Keyang Ye, Qiming Hou, and Kun Zhou.
\newblock 3d gaussian splatting with deferred reflection.
\newblock In \emph{ACM SIGGRAPH}, 2024.

\bibitem[Yifan et~al.(2019)Yifan, Serena, Wu, {\"O}ztireli, and Sorkine-Hornung]{yifan2019differentiable}
Wang Yifan, Felice Serena, Shihao Wu, Cengiz {\"O}ztireli, and Olga Sorkine-Hornung.
\newblock Differentiable surface splatting for point-based geometry processing.
\newblock \emph{ACM TOG}, 38\penalty0 (6):\penalty0 1--14, 2019.

\bibitem[Zhang et~al.(2021{\natexlab{a}})Zhang, Luan, Wang, Bala, and Snavely]{zhang2021physg}
Kai Zhang, Fujun Luan, Qianqian Wang, Kavita Bala, and Noah Snavely.
\newblock Physg: Inverse rendering with spherical gaussians for physics-based material editing and relighting.
\newblock In \emph{CVPR}, 2021{\natexlab{a}}.

\bibitem[Zhang et~al.(2018)Zhang, Isola, Efros, Shechtman, and Wang]{zhang2018unreasonable}
Richard Zhang, Phillip Isola, Alexei~A Efros, Eli Shechtman, and Oliver Wang.
\newblock The unreasonable effectiveness of deep features as a perceptual metric.
\newblock In \emph{CVPR}, 2018.

\bibitem[Zhang et~al.(2021{\natexlab{b}})Zhang, Srinivasan, Deng, Debevec, Freeman, and Barron]{zhang2021nerfactor}
Xiuming Zhang, Pratul~P Srinivasan, Boyang Deng, Paul Debevec, William~T Freeman, and Jonathan~T Barron.
\newblock Nerfactor: Neural factorization of shape and reflectance under an unknown illumination.
\newblock \emph{ACM TOG}, 40\penalty0 (6):\penalty0 1--18, 2021{\natexlab{b}}.

\bibitem[Zhang et~al.(2022)Zhang, Sun, He, Fu, Jia, and Zhou]{zhang2022modeling}
Yuanqing Zhang, Jiaming Sun, Xingyi He, Huan Fu, Rongfei Jia, and Xiaowei Zhou.
\newblock Modeling indirect illumination for inverse rendering.
\newblock In \emph{CVPR}, 2022.

\end{thebibliography}
}
\clearpage
\maketitlesupplementary
\setcounter{section}{0}
\setcounter{figure}{0}
\renewcommand{\thetable}{\Alph{table}}
\renewcommand{\thefigure}{\Alph{figure}}
\renewcommand{\thesection}{\Alph{section}}

In the supplementary material, we first provide detailed explanations of the HDR processing steps in ~\cref{supp:processing}. Next, we illustrate the effects of the proposed shading function in \cref{supp:shading}. Additionally, object-wise comparison and extended  qualitative comparisons are presented in \cref{supp:obj_wise} and \cref{supp:visual}. 

\section{HDR Processing for Relighting}
\label{supp:processing}
In Sec. 3.4, we discussed how various design choices influence the interpretation of the learned environment maps and how novel HDR maps should be processed to align with this interpretation for relighting. \cref{fig:env_int} summarizes the four options, with the red dotted block indicating the processing applied to standard HDR maps to obtain the relighting results presented in Tab. 1 of the main text. While omitting gamma correction during training is theoretically inconsistent with a linear light transport, it is included for completeness and to reflect common practice where it is often overlooked.
\begin{figure}[h]
    \centering
    \vspace{-1em}
    \includegraphics[width=1.0\linewidth]{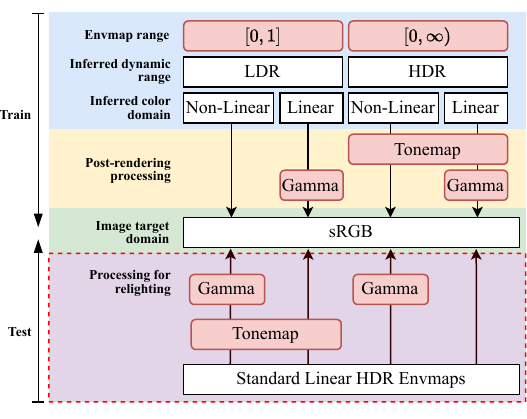}
    \caption{Design choices and relevant implication on the learned environment maps. During relighting (highlighted by the red dotted block), the input HDR map are processed to match the interpretation.}
    \vspace{-1em}
    \label{fig:env_int}
\end{figure}

Since GShader~\cite{jiang2024gaussianshader} and 3DGS-DR~\cite{ye20243d} do not provide official relighting code, \cref{supp:gshader} and \cref{supp:gsdr} explain the processing applied to obtain their relighting results.
\subsection{GShader}
\label{supp:gshader}
GShader~\cite{jiang2024gaussianshader} constrains its learned environment maps to the range $[0, 1]$ and adopts a $y$-up coordinate convention for light querying. As illustrated in \cref{fig:gshader_proc}, we perform coordinate transformations and clipping on latitude-longitude HDR environment maps for relighting. Without clipping, the rendered object appears overexposed.
\begin{figure}[t]
    \centering
\includegraphics[width=1.0\linewidth]{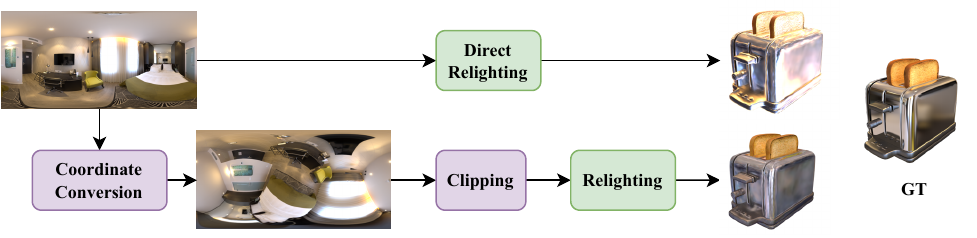}
    \caption{Environment map processing for GShader.}
    \vspace{-1em}
    \label{fig:gshader_proc}
\end{figure}
\subsection{3DGS-DR}
\label{supp:gsdr}
The learned environments in 3DGS-DR~\cite{ye20243d} are not range-limited and adopts a $y$-down coordinate
system. Since the official implementation visualizes learned environment maps with a sigmoid function, we use an inverse sigmoid to adjust the lighting values. As shown in~\cref{fig:gsdr_proc}, we apply coordinate transformation, clipping and inverse sigmoid on the HDR maps for relighting. Without the adjustment, the relighting output appears dull, and the highlight pattern also shifts incorrectly (from the upper surface to the front surface) due to the mismatched coordinate system.
\begin{figure}[t]
    \centering
\includegraphics[width=1.0\linewidth]{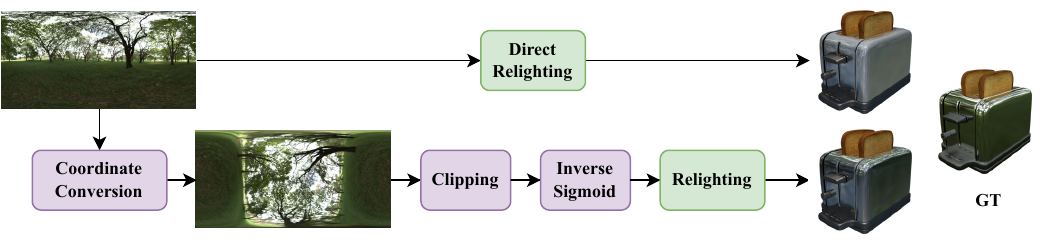}
    \caption{Environment map processing for 3DGS-DR.}
    \vspace{-1em}
    \label{fig:gsdr_proc}
\end{figure}
\section{Visual Ablation of the Shading Function}
\label{supp:shading}
The proposed shading function offers a more flexible material representation, which proves particularly advantageous for specular surfaces as shown in~\cref{fig:shading_ablate}. Although relighting differences are less pronounced for diffuse objects, training by the original shading function tends to misclassify non-metallic surfaces, such as the ficus leaves.
\begin{figure}[hb]
    \centering
\includegraphics[width=0.95\linewidth]{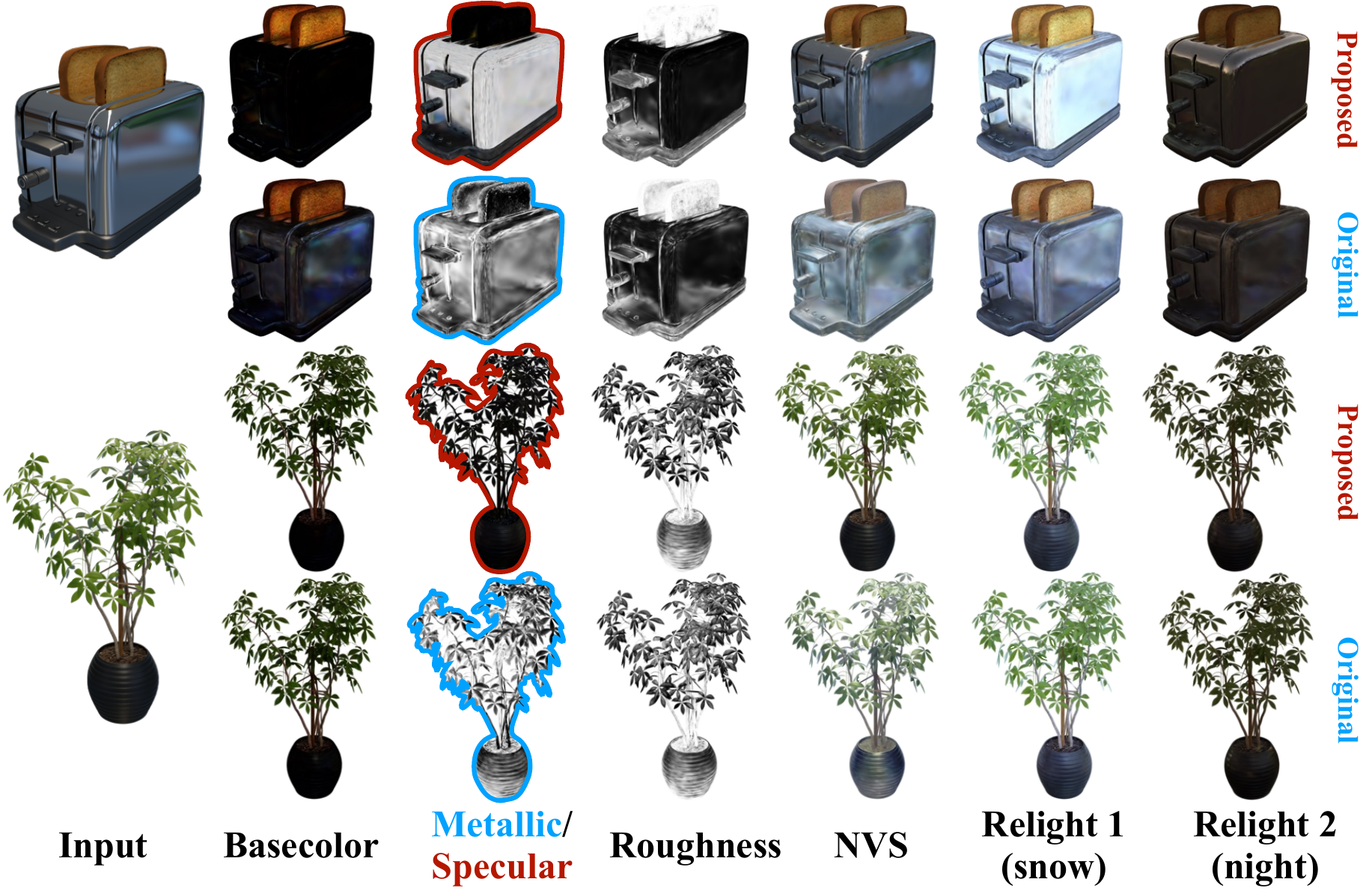}
    \caption{Comparison of the original and the proposed shading function.}
    \label{fig:shading_ablate}
\end{figure}

\begin{table*}[ht]
    \footnotesize
    \setlength\tabcolsep{5pt}
    \centering
    \caption{\fboxsep0.75pt Object-wise relighting performance. Methods optimized over single-env (\scalebox{0.75}{\ding{70}}) and dual-env (\scalebox{0.75}{\ding{71}}) inputs differ in the choice of training split B. Best and second best results are highlighted in \textbf{\colorbox{lighttop1}{bold}} and \textit{\colorbox{lighttop2}{italic}} respectively.}
    \vspace{-1em}
    \begin{tabularx}{0.98\linewidth}{X*{13}{c}}
        \hline

        \hline

        \hline
        &\multicolumn{8}{c}{\textit{Relit NeRF Synthetic}}&\multicolumn{5}{c}{\textit{Relit Shiny Blender}} \\
        \cmidrule(r){2-9}\cmidrule(r){10-14}
        &\textbf{chair}&\textbf{drums}&\textbf{ficus}&\textbf{hotdog}&\textbf{lego}&\textbf{ship}&\textbf{material}&\textbf{mic}&\textbf{helmet}&\textbf{coffee}&\textbf{musclecar}&\textbf{toaster}&\textbf{teapot}
        \\
	\hline
	
	\hline
	\multicolumn{14}{c}{PSNR$\uparrow$}\\
	\hline
	3DGS-DR\textsuperscript{\ding{70}}\cite{ye20243d}&21.25&21.82&25.78&19.34&21.38&19.06&20.01&26.42&20.33&18.09&19.62&17.35&25.46
	\\
	GS-IR\textsuperscript{\ding{70}}\cite{liang2024gs}&20.56&22.53&\cellcolor{lighttop2}\textit{27.45}&22.22&24.00&\cellcolor{lighttop2}\textit{21.79}&21.38&27.58&19.63&20.40&22.93&16.51&26.44
	\\
	R3DG\textsuperscript{\ding{70}}\cite{R3DG2023}&23.67&21.23&25.60&24.38&24.13&20.53&21.55&26.77&21.17&17.66&22.47&16.66&24.54
	\\
	GShader\textsuperscript{\ding{70}}\cite{jiang2024gaussianshader}&20.83&19.69&25.56&18.80&16.42&17.37&20.03&26.16&19.07&18.35&21.19&18.55&24.03
	\\
	ReCap\textsuperscript{\ding{70}}~(ours)&21.49&\cellcolor{lighttop2}\textit{24.02}&24.82&22.43&23.22&20.08&\cellcolor{lighttop2}\textit{22.69}&29.70&\cellcolor{lighttop2}\textit{23.41}&18.38&25.74&18.89&27.59
	\\
	\hline
	3DGS-DR\textsuperscript{\ding{71}}\cite{ye20243d}&21.86&22.09&25.55&21.13&22.71&20.41&20.73&26.60&18.54&20.44&21.66&15.14&26.33
	\\
	GS-IR\textsuperscript{\ding{71}}\cite{liang2024gs}&21.18&22.70&27.21&21.74&24.07&21.51&21.80&26.84&19.41&20.27&22.74&16.05&26.34
	\\
	R3DG\textsuperscript{\ding{71}}\cite{R3DG2023}&24.08&21.25&25.17&23.57&24.50&18.34&22.17&25.76&20.74&18.66&22.09&17.01&25.33
	\\
	GShader\textsuperscript{\ding{71}}\cite{jiang2024gaussianshader}&21.33&21.11&25.71&18.70&18.47&18.98&20.98&27.17&19.51&18.28&22.63&18.06&24.24
	\\
	\hline
	TensoIR\textsuperscript{\ding{71}}\cite{jin2023tensoir}&\cellcolor{lighttop2}\textit{26.32}&20.77&24.29&\cellcolor{lighttop1}\textbf{26.06}&\cellcolor{lighttop2}\textit{27.08}&21.01&20.69&\cellcolor{lighttop2}\textit{29.77}&20.11&\cellcolor{lighttop1}\textbf{25.13}&\cellcolor{lighttop2}\textit{25.83}&15.43&\cellcolor{lighttop1}\textbf{35.89}
	\\
	TensoIR\textsuperscript{\ding{71}}\textsubscript{no scale}\cite{jin2023tensoir}&20.25&22.99&24.28&\cellcolor{lighttop2}\textit{24.50}&\cellcolor{lighttop1}\textbf{28.23}&\cellcolor{lighttop1}\textbf{21.98}&19.96&28.40&18.76&\cellcolor{lighttop2}\textit{23.66}&24.59&15.02&27.76
	\\
	GShader\textsuperscript{\ding{71}}\cite{jiang2024gaussianshader}~+~ours&23.25&21.06&26.63&20.24&19.17&18.80&19.97&27.18&19.81&20.34&21.84&\cellcolor{lighttop2}\textit{20.53}&24.20
	\\
	ReCap\textsuperscript{\ding{71}}~(ours)&\cellcolor{lighttop1}\textbf{26.63}&\cellcolor{lighttop1}\textbf{24.77}&\cellcolor{lighttop1}\textbf{28.24}&23.25&24.04&20.12&\cellcolor{lighttop1}\textbf{25.83}&\cellcolor{lighttop1}\textbf{31.31}&\cellcolor{lighttop1}\textbf{23.55}&22.17&\cellcolor{lighttop1}\textbf{28.99}&\cellcolor{lighttop1}\textbf{25.05}&\cellcolor{lighttop2}\textit{31.65}
	\\
	\hline
	
	\hline
	\multicolumn{14}{c}{SSIM$\uparrow$}\\
	\hline
	3DGS-DR\textsuperscript{\ding{70}}\cite{ye20243d}&0.920&0.911&0.958&0.882&0.867&0.771&0.875&0.956&0.885&0.903&0.882&0.824&0.974
	\\
	GS-IR\textsuperscript{\ding{70}}\cite{liang2024gs}&0.897&0.901&0.956&0.889&0.874&0.790&0.864&0.954&0.849&0.896&0.895&0.767&0.969
	\\
	R3DG\textsuperscript{\ding{70}}\cite{R3DG2023}&0.924&0.895&0.952&0.904&0.880&0.774&0.864&0.955&0.865&0.881&0.899&0.784&0.969
	\\
	GShader\textsuperscript{\ding{70}}\cite{jiang2024gaussianshader}&0.914&0.891&0.951&0.885&0.821&0.777&0.871&0.956&0.889&0.908&0.896&0.824&0.973
	\\
	ReCap\textsuperscript{\ding{70}}~(ours)&0.922&\cellcolor{lighttop2}\textit{0.928}&\cellcolor{lighttop2}\textit{0.961}&0.900&0.887&0.801&\cellcolor{lighttop2}\textit{0.889}&0.970&\cellcolor{lighttop2}\textit{0.909}&0.910&\cellcolor{lighttop2}\textit{0.928}&0.824&\cellcolor{lighttop2}\textit{0.982}
	\\
	\hline
	3DGS-DR\textsuperscript{\ding{71}}\cite{ye20243d}&0.925&0.916&0.957&0.897&0.884&\cellcolor{lighttop2}\textit{0.804}&0.876&0.957&0.862&0.919&0.896&0.781&0.976
	\\
	GS-IR\textsuperscript{\ding{71}}\cite{liang2024gs}&0.900&0.901&0.954&0.885&0.868&0.793&0.861&0.950&0.848&0.894&0.888&0.756&0.966
	\\
	R3DG\textsuperscript{\ding{71}}\cite{R3DG2023}&0.931&0.897&0.951&0.904&0.883&0.761&0.871&0.952&0.867&0.891&0.899&0.792&0.972
	\\
	GShader\textsuperscript{\ding{71}}\cite{jiang2024gaussianshader}&0.920&0.901&0.954&0.892&0.843&0.794&0.878&0.961&0.889&0.909&0.911&0.820&0.974
	\\
	\hline
	TensoIR\textsuperscript{\ding{71}}\cite{jin2023tensoir}&\cellcolor{lighttop2}\textit{0.939}&0.877&0.946&0.911&\cellcolor{lighttop2}\textit{0.922}&0.777&0.862&\cellcolor{lighttop2}\textit{0.970}&0.825&0.891&0.910&0.684&0.978
	\\
	TensoIR\textsuperscript{\ding{71}}\textsubscript{no scale}\cite{jin2023tensoir}&0.920&0.893&0.946&\cellcolor{lighttop2}\textit{0.912}&\cellcolor{lighttop1}\textbf{0.925}&0.800&0.870&0.967&0.818&0.892&0.901&0.688&0.977
	\\
	GShader\textsuperscript{\ding{71}}\cite{jiang2024gaussianshader}~+~ours&0.935&0.905&0.960&0.900&0.855&0.795&0.888&0.963&0.903&\cellcolor{lighttop2}\textit{0.927}&0.906&\cellcolor{lighttop2}\textit{0.869}&0.975
	\\
	ReCap\textsuperscript{\ding{71}}~(ours)&\cellcolor{lighttop1}\textbf{0.951}&\cellcolor{lighttop1}\textbf{0.937}&\cellcolor{lighttop1}\textbf{0.972}&\cellcolor{lighttop1}\textbf{0.912}&0.908&\cellcolor{lighttop1}\textbf{0.815}&\cellcolor{lighttop1}\textbf{0.921}&\cellcolor{lighttop1}\textbf{0.975}&\cellcolor{lighttop1}\textbf{0.920}&\cellcolor{lighttop1}\textbf{0.940}&\cellcolor{lighttop1}\textbf{0.943}&\cellcolor{lighttop1}\textbf{0.910}&\cellcolor{lighttop1}\textbf{0.988}
	\\
	\hline
	
	\hline
	\multicolumn{14}{c}{LPIPS$\downarrow$}\\
	\hline
	3DGS-DR\textsuperscript{\ding{70}}\cite{ye20243d}&\cellcolor{lighttop2}\textit{0.055}&0.056&0.029&0.110&0.088&0.173&\cellcolor{lighttop2}\textit{0.081}&0.025&\cellcolor{lighttop2}\textit{0.116}&0.149&0.075&0.143&0.030
	\\
	GS-IR\textsuperscript{\ding{70}}\cite{liang2024gs}&0.072&0.069&0.036&0.121&0.090&0.185&0.094&0.032&0.176&0.152&0.080&0.221&0.037
	\\
	R3DG\textsuperscript{\ding{70}}\cite{R3DG2023}&0.060&0.068&0.037&0.105&0.088&0.188&0.103&0.035&0.171&0.152&0.078&0.210&0.035
	\\
	GShader\textsuperscript{\ding{70}}\cite{jiang2024gaussianshader}&0.078&0.068&0.038&0.133&0.119&0.192&0.089&0.028&0.134&0.158&0.075&0.157&0.037
	\\
	ReCap\textsuperscript{\ding{70}}~(ours)&0.060&\cellcolor{lighttop2}\textit{0.048}&\cellcolor{lighttop2}\textit{0.028}&0.103&\cellcolor{lighttop2}\textit{0.081}&\cellcolor{lighttop1}\textbf{0.162}&0.082&\cellcolor{lighttop2}\textit{0.019}&0.118&\cellcolor{lighttop2}\textit{0.140}&\cellcolor{lighttop2}\textit{0.057}&0.151&\cellcolor{lighttop2}\textit{0.023}
	\\
	\hline
	3DGS-DR\textsuperscript{\ding{71}}\cite{ye20243d}&0.057&0.055&0.032&0.102&0.085&0.187&0.087&0.028&0.148&0.150&0.073&0.185&0.031
	\\
	GS-IR\textsuperscript{\ding{71}}\cite{liang2024gs}&0.074&0.070&0.040&0.131&0.099&0.196&0.102&0.037&0.183&0.161&0.085&0.235&0.040
	\\
	R3DG\textsuperscript{\ding{71}}\cite{R3DG2023}&0.060&0.067&0.039&0.110&0.091&0.206&0.104&0.038&0.170&0.160&0.078&0.212&0.037
	\\
	GShader\textsuperscript{\ding{71}}\cite{jiang2024gaussianshader}&0.076&0.064&0.038&0.130&0.106&0.219&0.090&0.027&0.135&0.164&0.068&0.164&0.038
	\\
	\hline
	TensoIR\textsuperscript{\ding{71}}\cite{jin2023tensoir}&0.075&0.122&0.067&0.135&0.096&0.221&0.134&0.043&0.256&0.164&0.116&0.310&0.046
	\\
	TensoIR\textsuperscript{\ding{71}}\textsubscript{no scale}\cite{jin2023tensoir}&0.108&0.119&0.067&0.144&0.089&0.208&0.132&0.046&0.262&0.165&0.121&0.308&0.048
	\\
	GShader\textsuperscript{\ding{71}}\cite{jiang2024gaussianshader}~+~ours&0.063&0.062&0.032&\cellcolor{lighttop2}\textit{0.100}&0.095&0.205&0.088&0.024&0.125&0.146&0.071&\cellcolor{lighttop2}\textit{0.124}&0.038
	\\
	ReCap\textsuperscript{\ding{71}}~(ours)&\cellcolor{lighttop1}\textbf{0.042}&\cellcolor{lighttop1}\textbf{0.042}&\cellcolor{lighttop1}\textbf{0.023}&\cellcolor{lighttop1}\textbf{0.091}&\cellcolor{lighttop1}\textbf{0.068}&\cellcolor{lighttop2}\textit{0.165}&\cellcolor{lighttop1}\textbf{0.063}&\cellcolor{lighttop1}\textbf{0.016}&\cellcolor{lighttop1}\textbf{0.094}&\cellcolor{lighttop1}\textbf{0.120}&\cellcolor{lighttop1}\textbf{0.048}&\cellcolor{lighttop1}\textbf{0.088}&\cellcolor{lighttop1}\textbf{0.021}
	\\

    \hline

    \hline

    \hline
    \end{tabularx}

    \label{tab:obj_wise}
    \vspace{-1em}
\end{table*}
\section{Object-wise results}
\label{supp:obj_wise}
We present object-wise relighting performance in~\cref{tab:obj_wise}. Objects with strong self-shadowing/self-reflection (e.g., \text{hotdog}, \text{coffee}, \text{lego} as shown in~\cref{fig:fail_case}) are difficult for ReCap as we do not handle secondary rays. \text{Ship} poses a challenge for all methods with its translucent material requiring the modeling of subsurface scattering.
\section{Additional Results (\cref{fig:car} to \cref{fig:mic})}
\label{supp:visual}
\begin{figure}[h]
    \centering
    \vspace{-3em}
    \includegraphics[width=1.0\linewidth]{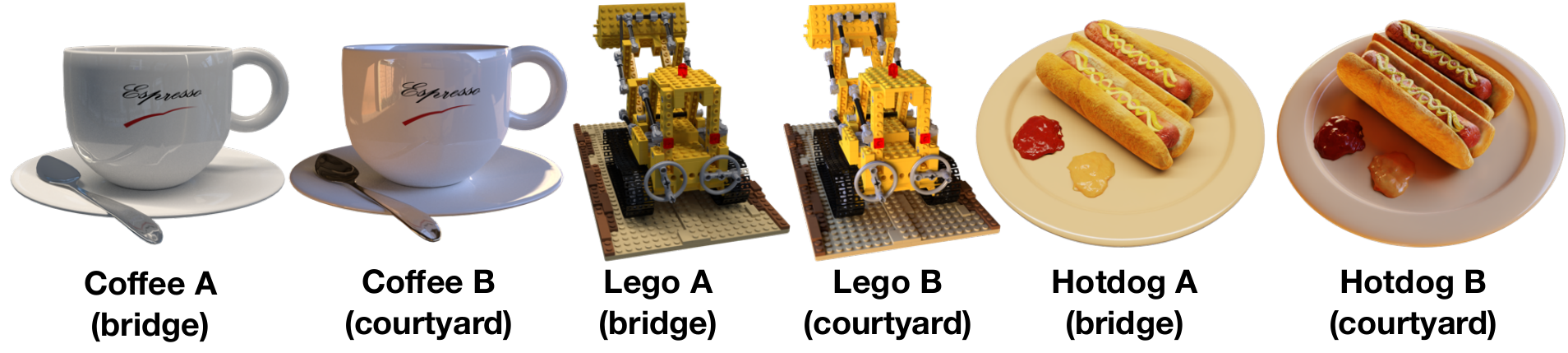}
    \vspace{-1em}
    \caption{Training views of objects with self-shadowing/self-reflection.}
    \label{fig:fail_case}
\end{figure}

\begin{figure*}[h]
    \centering
    \includegraphics[height=0.95\textheight]{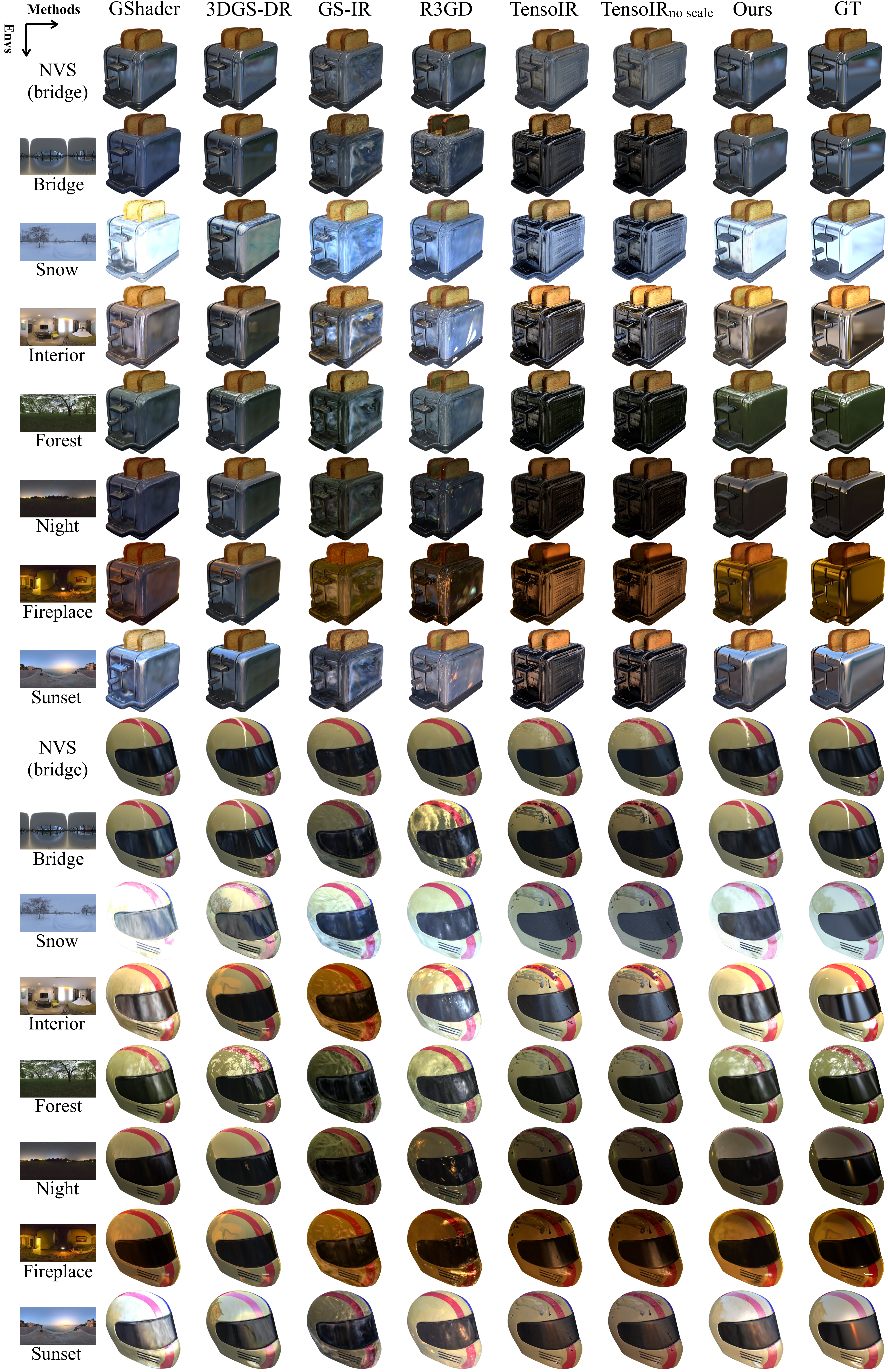}
    \vspace{-1em}
    \caption{Qualitative comparison of relighting results across different environment maps. Upper: toaster; lower: helmet.}
    \label{fig:car}
\end{figure*}

\begin{figure*}[h]
    \centering
    \includegraphics[height=0.95\textheight]{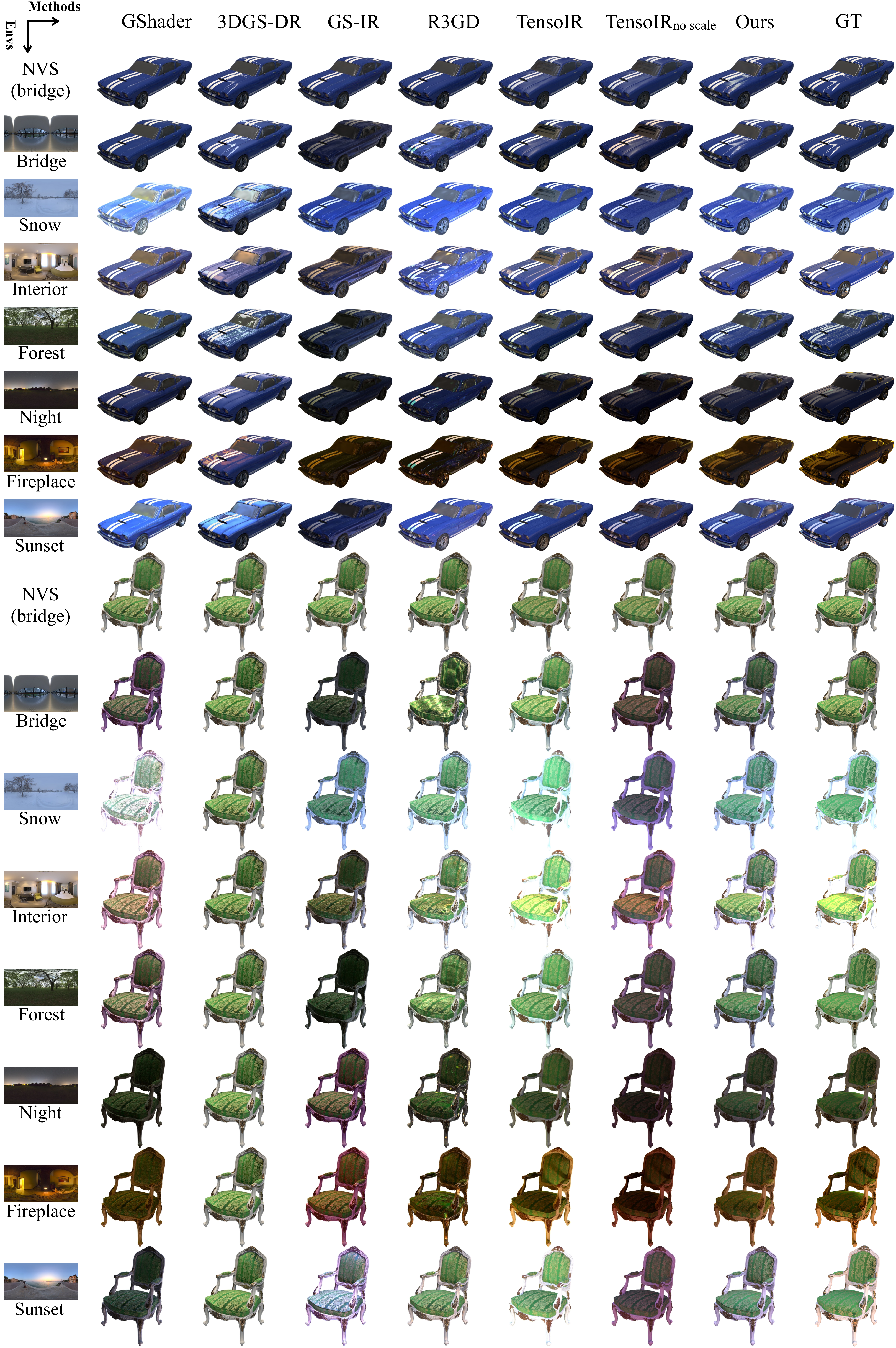}
    \vspace{-1em}
    \caption{Qualitative comparison of relighting results across different environment maps. Upper: musclecar; lower: chair.}
    \label{fig:toaster}
\end{figure*}

\begin{figure*}[h]
    \centering
    \includegraphics[height=0.95\textheight]{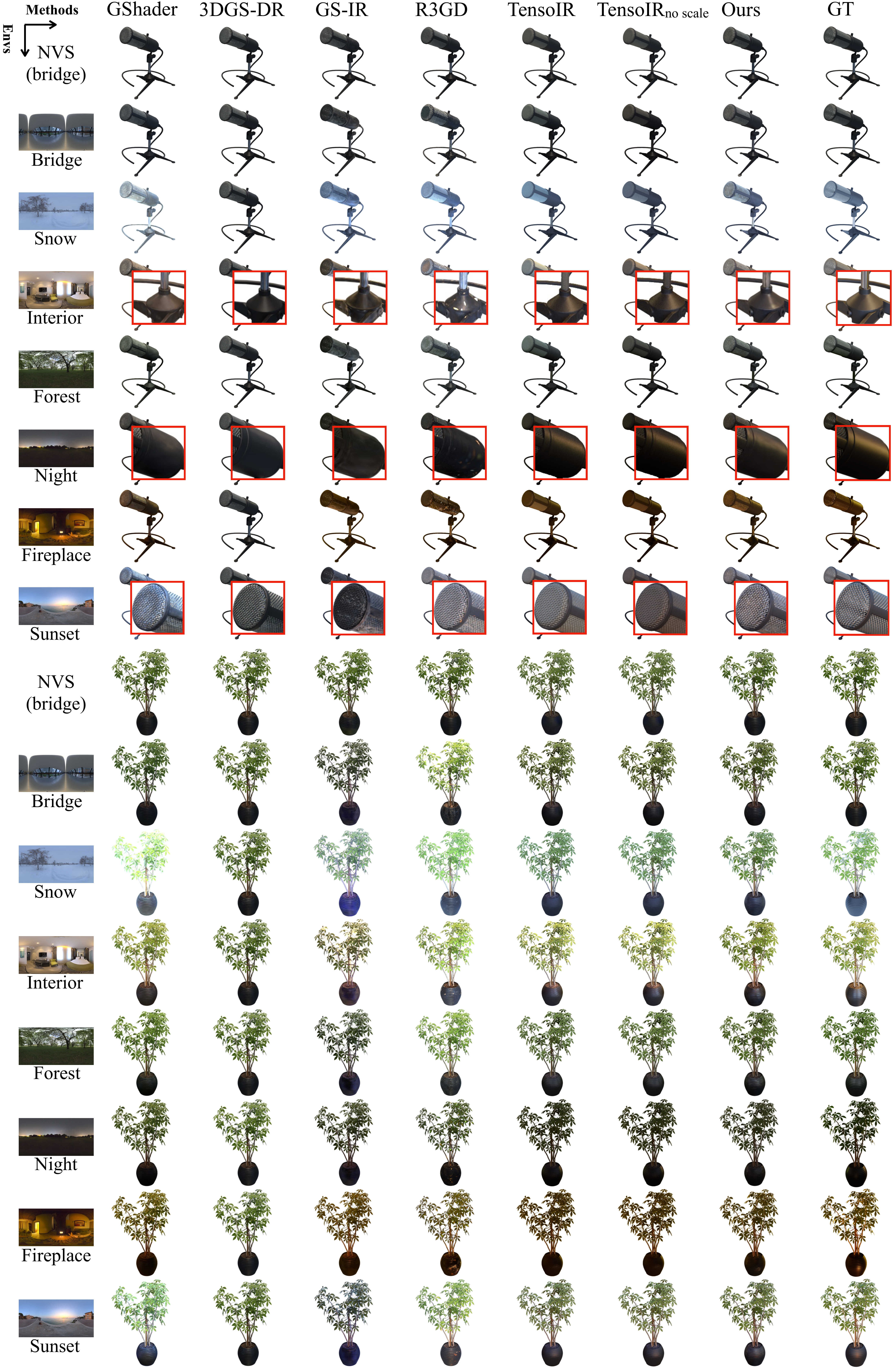}
    \vspace{-1em}
    \caption{Qualitative comparison of relighting results across different environment maps. Upper: mic; lower: ficus.}
    \label{fig:mic}
\end{figure*}
\newpage

\end{document}